\documentclass[a4paper,fleqn]{cas-sc}
\usepackage{caption}
\usepackage{natbib}
\usepackage{amsmath} 
\usepackage{url}
\usepackage{ragged2e}

\def\tsc#1{\csdef{#1}{\textsc{\lowercase{#1}}\xspace}}
\tsc{WGM}
\tsc{QE}
\tsc{EP}
\tsc{PMS}
\tsc{BEC}
\tsc{DE}

\ExplSyntaxOn
\keys_set:nn { stm / mktitle } { nologo }
\ExplSyntaxOff

\begin{document}
\captionsetup[figure]{labelfont=bf,labelformat=default,labelsep=period,name={Fig.}}
\let\WriteBookmarks\relax
\def\floatpagepagefraction{1}
\def\textpagefraction{.001}

\shorttitle{ }


\title [mode = title]{WikiLink: an encyclopedia-based semantic network for design innovation}




\author[2]{Haoyu Zuo}[type=editor,
                        auid=000,bioid=1,
                        orcid=0000-0003-3811-4479,style=chinese]


\ead{h.zuo19@imperial.ac.uk}




\author[1]{Qianzhi Jing}[style=chinese]
\ead{jingqz@zju.edu.cn}

\author[1]{Tianqi Song}[style=chinese]
\ead{holly1027@zju.edu.cn}

\author[1]{Huiting Liu}[style=chinese]
\ead{3190102536@zju.edu.cn}
\author[1]{Lingyun Sun}[style=chinese]
\ead{sunly@zju.edu.cn}

\author[2]{Peter Childs}
\ead{p.childs@imperial.ac.uk}

\author[1]{Liuqing Chen}[style=chinese]
\cormark[1]
\ead{chenlq@zju.edu.cn}





    
\address[1]{Department of Computer Science and Technology, Zhejiang University,Hangzhou,310030,China}

\address[2]{Dyson School of Design Engineering, Imperial College London,Exhibition Rd, South Kensington,SW7 2AZ,United Kingdom}



\cortext[cor1]{Corresponding author}




\begin{abstract}
\noindent Data-driven design and innovation is a process to reuse and provide valuable and useful information. However, existing  semantic networks  for design innovation is built on data source restricted to technological and scientific information. Besides, existing studies build the edges of a semantic network only on either statistical or semantic relationships, which is less likely to make full use of the benefits from both types of relationships and discover implicit knowledge for design innovation. Therefore, we constructed WikiLink, a semantic network based on Wikipedia.  Combined weight which fuses both the statistic and semantic weights between concepts is introduced in WikiLink, and four algorithms are developed for inspiring new ideas. Evaluation experiments are undertaken and results show that the network is characterised by high coverage of terms, relationships and disciplines, which proves the network’s effectiveness and usefulness. Then a demonstration and case study results indicate that WikiLink can serve as an idea generation tool for innovation in conceptual design. The source code of WikiLink and the backend data are provided open-source for more users to explore and build on.
\end{abstract}






\maketitle

\section{Introduction}

Design is a ubiquitous process that occurs throughout a variety of fields. Conceptual design is the early stage of design where an initial idea is formulated  \citep{childs2013}. The progression of conceptual design development requires a designer to fully utilize their innovation capability and existing knowledge. In other words, the creative attributes of conceptual design depend highly on a designer's ability to master, apply, and utilize human-centred, scientific and technological knowledge according to the design problem to provoke design innovation. Researchers have utilized a large amount of imagery data or textual data available on the internet to provide design intuition for novel ideas. This imposes a heavy challenge \citep{hao2014knowledgemap} for designers on how to effectively discover and acquire pertinent knowledge and information to promote design innovation.



With the advent of big data, semantic networks can represent associations well between ontology-based knowledge, making it easier and more intuitive to discover implicit knowledge for design innovation. The highly diverse nature of design suggests that design innovation can benefit from a multiplicity of distinct data. However, existing semantic networks for design innovation are built on data sources restricted to technological and scientific knowledge. Besides, existing studies build the edges of a semantic network only on either statistical or semantic relationships, which is less likely to make full use of the benefits from both types of relationships and discover implicit knowledge for design innovation.

To address the challenges highlighted, this study proposed an encyclopedia based network called WikiLink for design innovation. The source code of WikiLink is published on https://github.com/zju-d3/WikiLink. The main contributions of this paper can be summarized as follows:
\begin{enumerate}[(1)]
\item A semantic network for design innovation is constructed. Wikipedia is applied as the data source for the semantic network, which contains information from a wide range of fields and expands the data to a new boundary.
\item A combined weight is introduced for the relationship in the semantic network. The combined weight mixes the statistical relationship and semantic relationship which better captures the implicit connection between concepts for design innovation. Four algorithms are further developed for design which enables the retrieval with different levels and manners.
\item The constructed semantic network for design innovation is further developed as a tool. An evaluation and demonstration for the tool are conducted subsequently, the results show that WikiLink can effectively provide design stimuli for idea generation.
\end{enumerate}

The paper is organised as follows: section 2 describes the state of knowledge and background for the research, and section 3 introduces the process of constructing WikiLink. Section 4 presents the experimentation including the results on coverage of concepts, coverage of relationships, coverage of disciplines and term to term relationships. Section 5 demonstrates the use of four functions in WikiLink and presents a design case with WikiLink. Finally, section 6 concludes with limitations and suggestions for further research directions.

\section{Related Work}

\subsection{Design innovation and idea generation}
Design can be regarded as the process of conceiving, developing and realising products, artefacts, processes, systems, services, platforms and experiences with the aim of fulfilling identified or perceived needs or desires typically working within defined or negotiated constraints~\citep{childs2013}. Design innovation is the progress of creating innovative design, which needs the designer to fully utilize their ability to generate a design idea. Normally, the whole design innovation process can benefit from considering as many ideas as possible \citep{liu2003towards}. Ideas, especially creative ideas, are an essential part of the design innovation process \citep{han2018combinator,han2018acomputational}. 



Much research has endeavored to propose novel approaches for idea generation. The diverse idea generation techniques include brainstorming \citep{osborn1953appliedimagination}, brainwriting \citep{geschka1983creativity}, checklists \citep{ivanov2014satisfacton}, and synectics \citep{vangundy1988techniques}. Recently, data-driven approaches have attracted researchers' attention. In the process of design innovation, data-driven approaches attempt to uncover useful design knowledge from huge, unstructured, heterogeneous, and highly contextualized data resources \citep{shi2017data,cheong2017automated,luo2021guiding}. Researchers emphasize the importance of generating creative ideas in design the innovation process from big data \citep{howard2008describing,kwon2018kwon} and further indicate that creative ideas can originate from diverse existing knowledge and defined associations.

\subsection{Semantic network}

A semantic network is a graph with nodes representing concepts or individual objects and edges representing relationships or associations among concepts \citep{sowa1987semantic}. The use of a semantic network can help integrate and migrate valuable, unstructured data into systematic robust knowledge for design innovation \citep{gorti1998anobject,rezgui2011pastpresent,georgiev2018enhancing}.

When design work is completed, a great number of data and information are usually accumulated and reported afterwards \citep{ackoff1989data}, in the format of proceedings, literature, patents or public reports. These pieces of recorded information are expected to be transformed into design knowledge, which is expected to be reused for unhappened design tasks, to speed up more design work. When considering knowledge reuse, common knowledge sources generally include research papers, patent documents, encyclopedias.

Academic papers and patents are original research outcome or totally new inventions, which contain rich scientific and technological knowledge. Several attempts \citep{munoz2016modeling,fu2013discovering,he2019mining,mccaffrey2018approach,shi2017data,sarica2020technet} have been made to apply the academic paper and patents to a design innovation task. However, one of the major limitations is that patents and scientific literature are restricted to only technological and scientific knowledge \citep{shibata2008detecting,furukawa2015identifying,li2019forecasting,ernst2003patent}, while the nature of design tasks is of high diversity and complexity, with broad coverage of disciplines. To address the issue, an encyclopedia can be applied for design innovation since the most notable advantage of an encyclopedia is that it contains information from a wide range of fields and can expand the design knowledge coverage to a wider boundary compared with paper and patents \citep{kwon2018toward}.


\subsection{Semantic network for design innovation}

Over the past decade, several general semantic networks have been developed such as the lexical database WordNet \citep{fellbaum2010wordnet} and ConceptNet \citep{speer2017conceptnet}. These general semantic networks were first developed for artificial intelligence tasks such as machine translation, natural language understanding \citep{sowa1987semantic}.  These lexical semantic networks are utilized increasingly in the engineering design domain. They are often employed as the backend knowledge to computational tools for design idea generation and analysis \citep{han2018combinator,hanadatadrivenapporach,bae2020spinneret,georgiev2018enhancing}.  However, these lexical databases built on common-sense knowledge are not specifically aimed at use in design innovation.

Thus, there is an impetus for developing a design innovation-focused semantic network to meet the growing demands for engineering knowledge discovery, technology information retrieval, engineering design aids and innovation management. An innovation-focused semantic network normally builds nodes retrieved from a reliable data source and establishes the association based on statistical or semantic relationship. The statistical relationship that represents the value on associations are assigned with a statistical calculation. For example, \cite{shi2017data} created a large semantic network with statistical relationships in the engineering and design domain. Its statistical relationships are built on the co-occurrence between each pair of words in nearly one million engineering papers and one thousand design posts. \cite{he2019miningandrepersent} created a semantic network with a core-periphery structure according to the word clouds embedding co-occurrences information. In this way, the semantic network built the edges on a statistical level and could support engineering and technology innovation from a statistical perspective.

The semantic relationships are the associations that there exist between the meanings of words and are applied in many design activities, such as analogy and metaphor methods \citep{johnson1992metaphor,goel1997design}. As for semantic network for design innovation, \cite{sarica2019technology} built a large-scale comprehensive semantic network of technology-related data for engineering knowledge discovery (TechNet). The semantic relationships between words are established by using natural language processing techniques to derive the vector of such terms. \cite{kim2012causality} suggest a cause-and-effect relationship to build a cause-and-effect function network to support technology innovation. With semantic relationships, the network could support data integration, knowledge discovery and in-depth analysis from a semantic perspective \citep{,sarica2021designknolwedge, sarica2019engineeringknowledge,sarica2021idea}.

This study plan to build a large encyclopedia based semantic network with statistical-semantic fused relationships. Inspired by the use of statistical relationship in a semantic network and semantic relationships in the design engineering domain, we aim to build a semantic network that combines the benefits of both the statistical relationship and the semantic relationship to better capture the implicit connection of cross-domain concepts to better stimulate design innovation.


\section{Construction of WikiLink}


In this section, we constructed WikiLink, a semantic network based on Wikipedia data. The Wikipedia items are regarded as the nodes, the interlinks between the items on the same page are regarded as the directly connected relationship (edges) between nodes. The edges in the network are assigned with a fused weight consisting of two types of weight, and four algorithms are proposed to retrieve relevant knowledge concepts and relationships for design innovation.

\subsection{Data source}
While patents and scientific literature focus on technological and scientific knowledge, an encyclopedia is an integrated source of general knowledge and specific knowledge, with broad coverage of disciplines. Wikipedia, as an online encyclopedia, is unrestricted by the weight and volume, and has the potential to be truly comprehensive in knowledge. Wikipedia is written and maintained by a community of volunteers and offers copies of available content to anyone to download. WikiLink processes on English Wikipedia pages before 3rd January 2021, comprised 6,408,679 articles. For each Wikipedia article, WikiLink extracts the titles, main text, "see also" and categories for further analysis. Figure \ref{fig:examplepage} is an example page of a Wikipedia article, containing a title, main text, "see also" and categories. It should be noted that articles with a colon in the title are excluded. These articles with a colon account for 10\% of total articles, which are Wikipedia's administrative pages and are not relevant as the core source of design information.

\subsection{Extraction process}
Wikipedia covers 13 main categories to group pages on similar subjects, with each main category having up to 6 layers of subcategories. The deeper the subcategory is, the more specific Wikipedia’s title will be. The articles are firstly filtered based on the indicated categories on their article pages to avoid too specific articles: only the articles within 3-layer subcategories are kept. The network is constructed based on these selected articles' title, main text, and "see also".

There are two parts in a semantic network: the nodes and relationships between. The nodes are from three sections in each Wikipedia article: the title, the hyperlinks in the main text, and the hyperlinks in the "see also" section. These hyperlinks in the main text are chosen as nodes since they are verified concepts in Wikipedia and indicate explicit associations between concepts as they occur with other concepts in the same articles. 

\begin{figure*}[t!]
	\centering
		\includegraphics[scale=0.7]{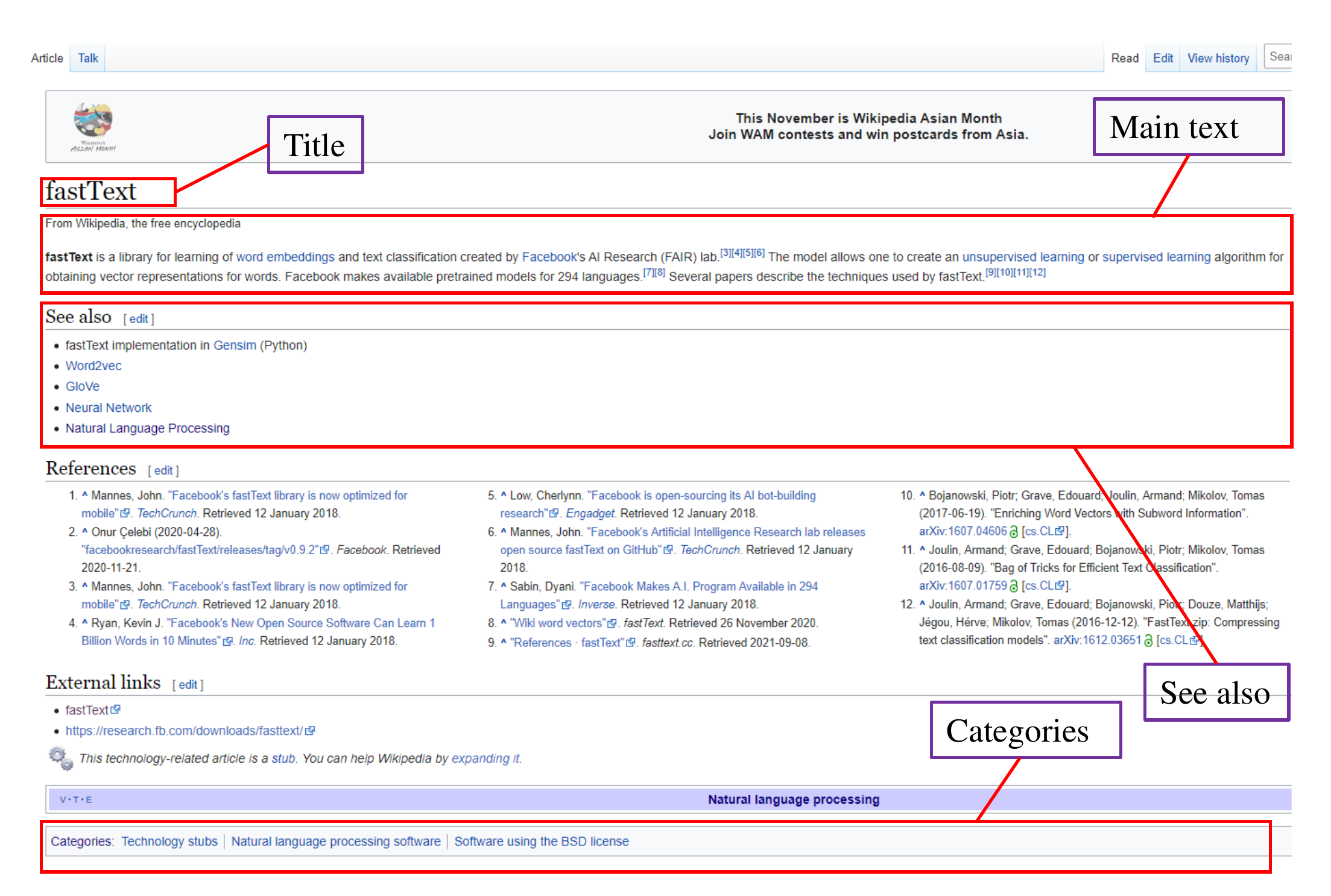}
	\caption{An example page in Wikipedia}
	\label{fig:examplepage}
\end{figure*}

The relationships are assumed to be established between two concepts if they co-occur in the same article. Two different criteria are applied for the raw weight accumulation of each relationship: since there is a large number of concepts in the main text, if two concepts co-occur in the main text, the weight is assigned a lower value to avoid dominant concepts; the concept in "see also" are intrinsically strong associations but with less amount compared with the concepts in the main text which is assigned a higher weight. The choice of different weight assignment is determined based on experimental results: if two concepts co-occur in the main text, the weight will be added with one; if two concepts occur in the "see also", the weight will be added with nine. The raw weight is accumulated and stored for later filtering. In this way, the nodes appearing in one article will be interlinked. Taking the content in Figure \ref{fig:examplepage} as an example, the nodes are “fastText”, “word embeddings” “Facebook” “unsupervised learning”, “supervised learning”, “Word2vec”, “Glove”, “Neural Network”, “Natural Language Processing”. The relationships are established between each pair of nodes because they co-occur in the same article. In this way, a network can be constructed by processing all articles in Wikipedia’s database.

\subsection{Construction of edge weights}

After the extraction process, an initial network with nodes and edges can be constructed. In the semantic network, explicit knowledge associations are direct edges linking pairs of nodes, and implicit knowledge associations are paths consisting of multiple edges, which means an implicit knowledge association is essentially a concatenation of a series of interconnected explicit knowledge associations \citep{shi2017data}. To evaluate the correlation degree of implicit knowledge associations, the weight of explicit knowledge associations should be quantified.


\subsubsection{Semantic cosine similarity weight}

In the construction process, the explicit associations are built based on the interlinked concepts within pages and the corresponding raw weights are statistically calculated. These statistical relationships construct the basic edges in a semantic network from a statistical perspective, which provides the foundation for WikiLink and statistical intuition for information retrieval. While in design activities, the semantic relationship also contributes much to design innovation such as analogy and metaphor methods \citep{hey2008analogies,linsey2012design} from a semantic perspective. Inspired by the implication of semantic relationship in design innovation activities, the statistical association between two concepts can be combined and balanced with the semantic similarity for boosting design innovation. The semantic similarity can be obtained by transforming all words to vectors and calculating the semantic cosine similarity between these vectorized concepts. The conventional word embedding methods like Word2Vec train a unique word embedding for every individual word. However, Wikipedia contains a large number of terms, with some of them even being new terms out of vocabulary. FastText \citep{bojanowski2017enriching,joulin2016fasttext} can solve this issue by treating each word as the aggregation of its subwords. The vector for a word is simply taken to be the sum of all vectors of its component char-ngrams. In this way, fastText can obtain vectors even for out-of-vocabulary (OOV) words, or the new terms in Wikipedia, by summing up vectors for its component char-ngrams, provided that at least one of the char-ngrams was present in the training data. When all concepts have been represented as word vectors, all edges connecting two nodes are assigned with a value by calculating the semantic cosine similarity between these vectors.

\subsubsection{Global normalization and local normalization}

In many design models, the design innovation process usually involves two important phases: divergence and convergence. For example, there are rounds of divergent and convergent phases in the "double diamond" design process model \citep{design_council_2017}. Divergence is a phase that encourages exploring different solutions as much as possible while convergence follows a particular set of logical steps to arrive at one solution which in some cases is a "correct" solution. Inspired by the principles of divergence and convergence, the retrieval behaviors can be facilitated in two distinct ways: a "general" and "specific" ways. "General" means the nodes are common and basic concepts with a relatively general meaning, which tends to lead divergent thinking in a design innovation process. While "specific" means the nodes are detailed and domain-specific concepts, which has higher potential to guide convergent thinking. The "general" and "specific" retrieval are realized by normalizing the raw weight with a globalization method as shown in equation (1) and a localization method as shown in equation (2):



\begin{equation}
w_{ij}^{g} = \left (w_{ij}-w_{min}  \right )/\left (w_{max}-w_{min}  \right )
\end{equation}

\begin{equation}
w_{ij}^{l} = w_{ij}/S_{i}
\end{equation}
where \textit{w}\textsubscript{\textit{max}} and \textit{w}\textsubscript{\textit{min}} are the maximum and minimum value of the raw weight in the whole network. \textit{w}\textsubscript{\textit{ij}} is the raw weight between the node i and node j, \textit{S}\textsubscript{\textit{i}} is the sum value of all raw weights of edges around node i. 

The global normalization performs feature scaling normalization from a global perspective, in which  
\textit{w}\textsubscript{\textit{ij}}\textsuperscript{\textit{g}}
expresses the significance of the strength compared to the whole network. Global normalization tends to retrieve more "general" concepts\citep{shi2017data}. The local normalization performs feature scaling normalization from a local perspective, in which \textit{w}\textsubscript{\textit{ij}}\textsuperscript{\textit{i}}
expresses the relative importance of the strength compared to its own adjacent value. Local normalization tends to extract more domain-specific concepts.

\subsubsection{Geometric mean and harmonic mean}
Since an implicit knowledge association is essentially a concatenation of a series of explicit associations, the accumulation of the strength of contained explicit associations (edges) can potentially indicate the correlation degree of the implicit association (path). Therefore, in order to reflect the overall strength of all the explicit associations in an arbitrary implicit association, the retrieval behaviors can be facilitated in two distinct ways: one type of retrieval, referred to as "basic", is a short implicit association across fewer edges focusing on relevant concepts which tend to be in the same domain while another type, referred to as "professional", is a long implicit association with more edges across multiple distant domains. Therefore, the geometric mean (GM) and the harmonic mean (HM) are applied on the normalized weights for different design innovation behaviors. 

The geometric mean(GM) and harmonic mean(HM) are given in equation (3) and (4) respectively:

\begin{equation}
\textrm{GM:} w_{(k_{1}-k_{2}-\cdots -k_{n+1})}=\sqrt[n]{\prod_{k=1}^{n}w_{k,k+1}}
\end{equation}

\begin{equation}
\textrm{HM:} w_{(k_{1}-k_{2}-\cdots -k_{n+1})}=\frac{n}{\sum_{k=1}^{n}\frac{1}{w_{k,k+1}} }
\end{equation}
where the \textit{w}\textsubscript{\textit{(k\textsubscript{1}-k\textsubscript{2}-...-k\textsubscript{n+1})}} is the overall weight of the path, \textit{w}\textsubscript{\textit{k,k+1}} is each weight along the path.

\subsection{Four algorithms for design innovation}

The primary use of the design semantic network is to retrieve relevant knowledge concepts and relationships for design innovation. In addition to retrieving around a single concept, retrieving the implicit associations between two distant knowledge concepts is also introduced. Four algorithms are developed by applying the normalization and mean methods to the proposed retrieval approach. The four algorithms, which are "Explore-General", "Explore-Specific", "Search Path-Basic" and "Search Path-Professional" are applied as four functions in WikiLink.

\begin{figure*}[t!]
	\centering
		\includegraphics[scale=0.45]{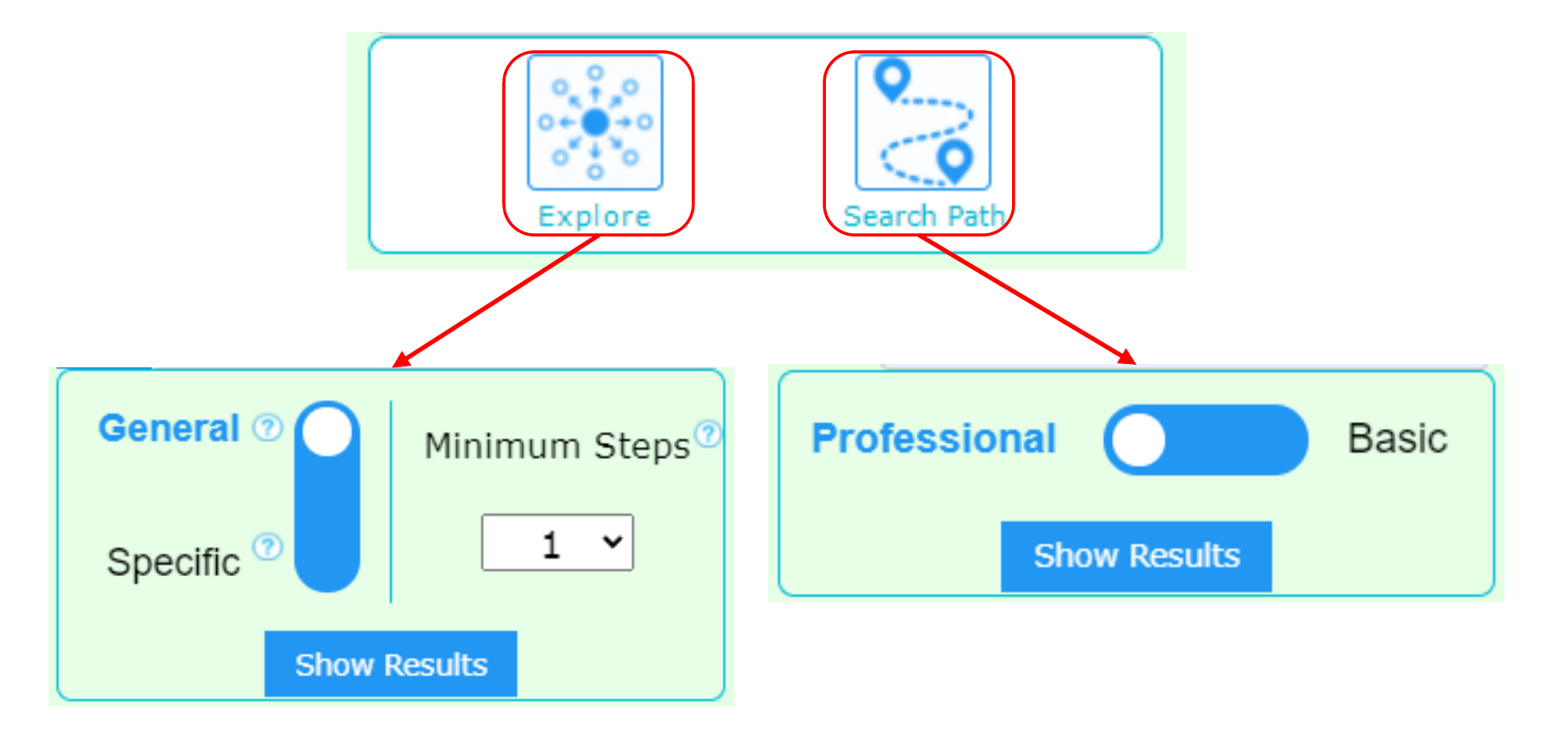}
	\caption{Four functions in the panel of WikiLink}
	\label{fig:fourfunctions}
\end{figure*}

The "Explore" algorithm is used to explore and retrieve around a single knowledge concept. The retrieved results can be classified as either "general" or "specific". The "Explore" function panel in WikiLink is shown in Figure \ref{fig:fourfunctions}. Specifically, since it is preferred to retrieve both "general" and "specific" knowledge concepts related to a query, we apply two different normalization algorithms with distinct retrieval behaviours in this "Explore" function. One is global normalization to retrieve "general" concepts for divergence, and the another is local normalization to retrieve "specific" concepts for convergence. The overall weight is calculated on a combination of the statistical weight and the semantic weight. The algorithm for "Explore-General", and "Explore-Specific" are given in equations (5) and (6) respectively:

\begin{equation}
w_{explore}^{general}=0.3\times (1-w_{semantic}^{}) + 0.7\times w_{}^{g} 
\end{equation}

\begin{equation}
\begin{split}
w_{explore}^{specific}=0.2\times (1-w_{semantic}^{}) + 0.8\times w_{}^{l} 
\end{split}
\end{equation}
where the \textit{w}\textsubscript{\textit{semantic}} is the semantic cosine similarity weight, \textit{w}\textsuperscript{\textit{g}}  is the statistical weight after global normalization and \textit{w}\textsuperscript{\textit{l}} is the statistical weight after local normalization.The weight in the algorithm are determined based on experimental results.

The "Explore" algorithms are further combined with the single source Dijkstra's shortest path algorithm, which starts from the source query to retrieve all reachable nodes in order from the shortest distance. In addition, a "Minimum Step" functionality is provided on the "Explore" panel, where knowledge associations with edges less than the number of the defined minimum step are filtering out for paths with fewer steps. Therefore, the knowledge associations are retrieved and ranked under the combined weight with the minimum step.

The "Search Path" algorithm is used to find implicit associations as paths are given two knowledge concepts. The retrieval result can be classified as either "basic" or "professional", where "basic" means the path is short and nodes are general concepts while "professional" means the paths are long and nodes are domain-specific concepts. The "Search Path" function panel in WikiLink is shown on the right side of Figure \ref{fig:fourfunctions}. Specifically, besides two different normalization algorithms, the geometric mean(GM) is further applied to retrieve short implicit associations across fewer edges focusing on relevant knowledge while harmonic mean(HM) is applied to retrieve long implicit associations with more edges across multiple domains. 

The algorithm of "Search Path-Basic" and "Search Path-Professional" are given in equations (7)(8)  and (9)(10) respectively:

\begin{equation}
w_{(k_{1}-k_{2}-\cdots -k_{n+1})}^{basic}=\sqrt[n]{\prod_{k=1}^{n}w_{k,k+1}^{basic}}
\end{equation}

\begin{equation}
\begin{split}
w_{k,k+1}^{basic}=0.3\times (1-w_{semantic}^{}) + 0.7\times w_{}^{g} 
\end{split}
\end{equation}

\begin{equation}
w_{(k_{1}-k_{2}-\cdots -k_{n+1})}^{professional}=\frac{n}{\sum_{k=1}^{n}\frac{1}{w_{k,k+1}^{professional}} }
\end{equation}

\begin{equation}
\begin{split}
w_{k,k+1}^{professional}=0.2\times (1-w_{semantic}^{}) + 0.8\times w_{}^{l} 
\end{split}
\end{equation}
where the \textit{w}\textsubscript{\textit{semantic}} is the semantic cosine similarity weight, \textit{w}\textsuperscript{\textit{g}}  is the statistical weight after global normalization and \textit{w}\textsuperscript{\textit{l}} is the statistical weight after local normalization.





\section{Evaluation}

In this section, we conduct four studies on WikiLink to demonstrate its effectiveness and usefulness. Some other semantic networks, which are publicly accepted or aiming for design innovation, are selected as benchmarks during the comparison, including B-link, WordNet and ConceptNet. The evaluation is conducted from four perspectives, i.e., coverage of concepts, coverage of relationships, coverage of disciplines, term-to-term evaluation and effectiveness of combined relationships to provide an overview of the strengths and weaknesses of WikiLink.

\subsection{Coverage of golden concepts}
In order to demonstrate the feasibility of WikiLink, golden concepts, which are composed of words and terms, are defined as the benchmark to evaluate WikiLink's term coverage. The golden concepts are collected manually within an online source Encyclopedia Britannica through several steps. Firstly featured concepts are obtained from its website.
There are several categories of topics available concerning different domains, including culture, science, and technology. By gathering these classified words and terms, it is ensured that the collected data contains interdisciplinary knowledge. The original data is refined afterward by removing uncommon expressions and standardizing their formats. The aim of this step is to assure the precision and impartiality of the following evaluations. Eventually, we obtain a list of 468 words and terms, covering knowledge in 8 domains, and part of the concepts are shown in Table \ref{tab:category_table}.

\begin{table}[]
    \caption{The overview of golden concepts}
    \label{tab:category_table}
    \begin{tabular}{lp{6cm}}
    \toprule
    Categories & Related concepts \\ 
    \midrule
    Animal      & bird, chordate, coral, insect, sea otter, ...\\
    Art         & acting, ballade, chinese literature, emmy award, film, ...\\
    Event       & american civil war, bronze age, cold war, french revolution, hurricane katrina, ...\\ 
    Place       & africa, anatolia, berlin, cape town, indonesia, ...\\
    Plant       & carnivorous plant, venus flytrap, ...\\
    Science     & atmosphere, brain, carbohydrate, chemistry, disease, ....\\
    Sports      & athletics, boxing, gymnastics, rugby, ...\\
    Technology  & airplane, bicycle, industry, radar, smartphone, supercomputer, ...\\
    Topic       & accident, architecture, buddhism, cbs corporation, democracy, ...\\
    \bottomrule
    \end{tabular}
\end{table}

With these golden concepts, we then evaluate how many concepts are contained in WikiLink. The retrieval rate $C_R$, as shown in equation (\ref{eq:c_r}), is applied as the metric of concept retrieval:
\begin{equation}
    \label{eq:c_r}
    C_R = \frac{n_C}{N_C}
\end{equation}
where $n_C$ means how many concepts are contained in the network, while $N_C$ represents the number of golden concepts, which is 468 is this case. 

WordNet and ConceptNet are used as two benchmarks for evaluation. It is observed that WordNet only contains 209 concepts, resulting in a low $C_R$ rate of 0.449. The specific $C_R$ values of different categories are shown separately in Table \ref{tab:re_table}, from which we notice that WikiLink gives the highest retrieval rate, indicating that our network has a wider coverage of concepts compared with the other tools considered.

\begin{table}[]
    \centering
    \caption{Retrieving results of golden concepts}\label{tab:re_table}
    \label{tab}
    \begin{tabular}{lp{1.5cm}p{1.5cm}p{1.5cm}}
    \toprule
    \text{Categories} & \text{WordNet} & \text{ConceptNet} & \text{WikiLink} \\
    \midrule
    \textbf{Total Rate CR} & 0.449 & 0.810 & \textbf{0.938} \\
    art & 0.386  & 0.818  & \textbf{0.841}  \\
    animal & \textbf{1.000}  & \textbf{1.000}  & \textbf{1.000}  \\
    event & 0.037  & 0.630  & \textbf{0.963}  \\
    place & 0.602  & \textbf{1.000}  & \textbf{1.000}  \\
    plant & 0.333  & \textbf{1.000}  & \textbf{1.000}  \\
    science & 0.631  & \textbf{0.954}  & \textbf{0.954}  \\
    sports & 0.652  & \textbf{0.957}  & 0.913  \\
    technology & 0.636  & 0.818  & \textbf{0.909}  \\
    topic & 0.287  & 0.638  & \textbf{0.920}  \\
    \bottomrule
    \end{tabular}
\end{table}

To be specific, our approach involves more concepts in most categories and achieves the highest retrieval rate. In comparison, WordNet shows overall weaknesses, due to its inadequacy in processing two-word terms. ConceptNet has decent performance in the fields of art, science, sports, and technology, but it lacks strengths in certain categories such as topics and events.

This result can be explained by the limitation of ConceptNet’s construction properties. Even though the data source of ConceptNet includes two-word terms, such as stained glass, chemical element, and mental disorder, these terms are mostly composed of one adjective and one noun. 
Except for names of countries and regions, seldom are two-noun terms involved in ConceptNet.
Based on our observation, plenty of concepts in those two categories, i.e., topics and events, are composed of more than one noun, e.g., teacher education, Paris agreement, and pacific crest trail, which are exactly situations that ConceptNet lacks solution to. 
This explains ConceptNet's low $C_R$ rate for those two categories.
In contrast, our approach can deal with various kinds of terms, which explains its overall high coverage. This high coverage of concepts can support design innovation with a large concept space.

\subsection{Coverage of golden relationships}

A list of golden relationships is selected from the data source as the evaluation benchmark to quantitatively evaluate the performance of relationship coverage. Similar to the construction process of WikiLink, we extracted concept relationships from Encyclopedia Britannica's spotlight articles. Only those which are composed of golden concepts are retained. We randomly picked 1000 concept pairs from the retained ones and defined as golden relationships.

Denoting golden relationships as set $H$, we compare the performance of WikiLink with other tools in terms of the coverage of golden relationships. In this process, we retrieve all relationships between golden concepts from each tool, and denote these retrieved relationships as set $V$. The evaluation metric is defined as follows:

\begin{align}
\begin{aligned}
 R = \frac{\lvert V \cap H\rvert}{\lvert H\rvert} 
\end{aligned}
\label{alg:ass}
\end{align}
where $R$ indicates the retrieving rate of relationships. WordNet and ConceptNet are chosen as benchmarks, and the results are shown in the Table \ref{tab:eva_table}.

\begin{table}[t!]
    \centering
    \caption{Evaluation results of golden relationships}
    \label{tab:eva_table}
    \begin{tabular}{lp{1.5cm}p{1.5cm}}
    \toprule
    \text{Categories} & \text{Count} & \text{R} \\
    \midrule
    WordNet & 15 & 0.015 \\
    ConceptNet & 170 & 0.170 \\
    WikiLink & \textbf{721} & \textbf{0.721} \\
    \bottomrule
    \end{tabular}
\end{table}

Specifically, 15 relationships are retrieved from WordNet, which belong to golden relationships, leading to a significantly low $R$ value of only 0.015. 
This retrieving rate can be explained by WordNet's data structure. 
To our knowledge, WordNet only retrieves specific relationships, including "synonyms", "sister terms", "hypernyms", and "hyponyms", between two concepts, which leads to its huge deficiency in context association and results in a low retrieving rate.

The web API of ConceptNet is used to retrieve concepts and relationships. 
It turns out that there are 170 relationships which are found in the golden relationships, resulting in a $R$ value of 0.170. 
The retrieving rate can be understood from two perspectives. 
ConceptNet's network contains more concepts than WordNet, which can be observed from its $C_R$ value. In addition, it provides richer explanations for "relationships". 
In other words, as well as "synonyms" and "hypernyms", ConceptNet is also able to retrieve "related terms" and "terms with this context" for an arbitrary single concept. 
These two reasons both contribute to its retrieving rate.

In the end, 721 relationships can be retrieved from the golden relationships within WikiLink.
This can be explained by its largest number of concepts, and the relationships in our approach are defined differently, i.e., they are established between concepts that are shown on the same pages.
To summarize, WikiLink achieves a retrieving rate of 0.721 and shows the best performance. This high retrieving rate of relationships builds enough associations which can potentially contribute to design innovation.

\subsection{Coverage of categories}
\begin{figure}[!htbp]
	\centering
	\includegraphics[width=0.55\textwidth]{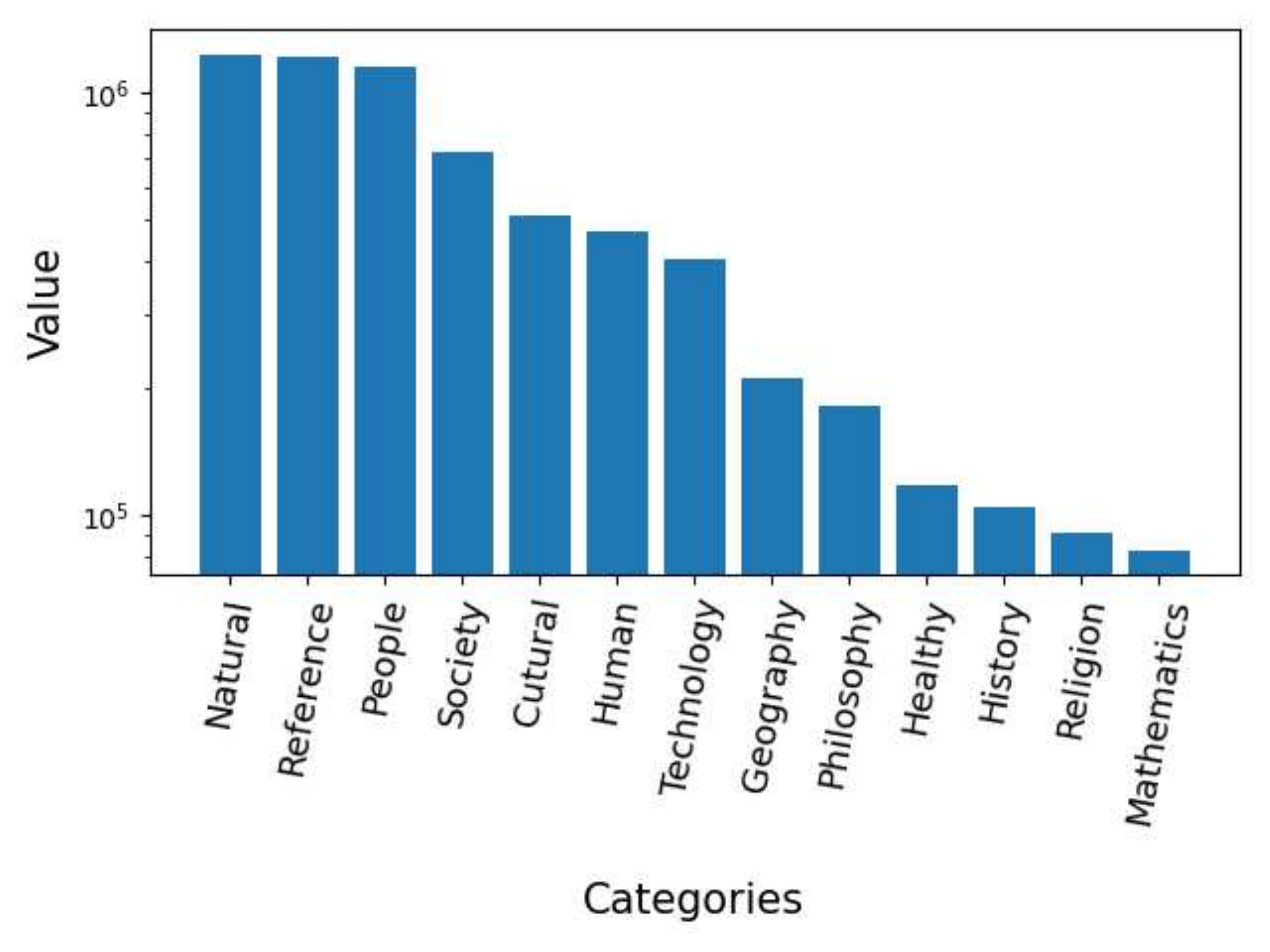}
	\caption{The distribution of concepts in WikiLink} 
	\label{fig:categories}
\end{figure}

To prove that WikiLink covers a wide range of categories, we categorize and count all the nodes in WikiLink according to Wikipedia's category rules. Wikipedia defines 13 main categories: cultural, geography, health, history, human, mathematics, natural, people, philosophy, religion, society, technology, and reference. By traversing all the items' categories in WikiLink, the distribution of the 13 categories is presented in Figure \ref{fig:categories}.

It can be seen from the graph that WikiLink's data have a wide distribution among 13 main categories, and the count of a particular main category can reach up to 100,000 level and even higher. Especially, the natural, people and reference categories have the largest counts, which are 1,241,491, 1,161,583 and 1,222,966 respectively. Rather than focusing only on technological and scientific knowledge, WikiLink is a more generic semantic network, with knowledge from a wide coverage of disciplines, which can be used in daily design innovation activities to obtain inspiration. Specifically, the data source of B-link mainly comes from scientific papers, which leads to the uneven distribution of each discipline, while WikiLink has a wide range of information in different fields and disciplines. Compared with TechNet, the result of WikiLink shows higher diversity as the distribution of TechNet is highly correlated with the distribution of patents, which may affect the inspiration of the design because of the coverage limitation, even though it contains a large number of domains within technology fields.

\subsection{Term to term evaluation}

To evaluate whether the computed edge weights are consistent with human judgement, thirty term pairs (three groups and each ten as a group) representing various degrees of relevance were prepared by language experts and ten students were employed to rate the relevance of each pair. The students scored semantic relevance and statistical relevance on a five-scale from one (not related) to five (highly related), and the average of scores is computed for each pair. The semantic relevance and statistical relevance are then combined as the weight in "Explore-General" algorithm. In this evaluation, only the "Explore-General" edge weights in the four algorithms is evaluated since the weight calculation in the four algorithms is all similar.

With the evaluation results, Cronbach’s alpha is used to measure the inter-rater reliability which is 0.78 as an acceptable result. Spearman's rank correlation coefficient is then used to assess the relationship between computed edge weights and human judgments. Table \ref{tbl1} shows the result of the Spearman rank correlation coefficients between the pairwise association values of the same term pairs. 
\begin{table}[t!]
\caption{Term to term evaluation results}\label{tbl1}
\begin{tabular}{@{}cc@{}}
\toprule
Group Number & Spearman Correlation \\ \midrule
1                & 0.69                 \\
2                & 0.89                 \\
3                & 0.64                 \\ \bottomrule
\end{tabular}
\end{table}

The hypothesis of Spearman correlation coefficient is then tested to determine whether the results are statistically significant. By checking the table of critical values, the three groups' Spearman’s rho are all greater than the critical value 0.57 (one tail,$\alpha$=0.05), so the null hypothesis is rejected. This proves that there is a strong correlation between the computed edge weights and human judgments , upheld by a significance level of \text{95\%}.


\subsection{Effectiveness of combined relationships}

As introduced in section 3, the statistical relationships between two concepts are established if they co-occur in the same article. Constructing the basic connection from a statistical perspective only could potentially lead to a phenomenon that the retrieval is dominated by some highly common concepts. These dominating common concepts decrease the retrieval probability of other useful concepts for design innovation. However, using semantic relationships only as the weight of edges is beneficial for design but might require in a longer association for implicit knowledge discovery. The semantic relationships are thus incorporated to balance the statistical relationship. To demonstrate the effectiveness of the proposed weight fusion, three types of retrieval results based on different relationships (networks with combined relationships, with statistical relationships, and with semantic relationships) are compared. The concept “health” is chosen for the “Explore” function and the concept pair “health \& 3d printing” is chosen for the “Search Path” function. 

\begin{table}[t!]

\caption{The high-correlated knowledge associations between "health" and "3d printing" with three different relationships}\label{Tlb:KA}
\begin{tabular}{p{1.4cm}<{}m{4cm}m{4.9cm}m{4cm}}
\toprule
~ & Combined relationship & Statistical relationship & Semantic relationship\\ \midrule
      Basic & \textbf{health} $\rightarrow$ economics $\rightarrow$ Massachusetts Institute of Technology $\rightarrow$ \textbf{3D printing} & \textbf{health} $\rightarrow$ education $\rightarrow$ United States $\rightarrow$ The New York Times $\rightarrow$ artificial intelligence $\rightarrow$  \textbf{3D printing} & \textbf{health} $\rightarrow$ health care $\rightarrow$ palliative care $\rightarrow$ intensive care unit $\rightarrow$ \textbf{3D printing} \\
      \hline
      Professional & \textbf{health} $\rightarrow$ construction $\rightarrow$ ladder $\rightarrow$ \textbf{3D printing} & \textbf{health} $\rightarrow$ physical fitness $\rightarrow$ physical strength $\rightarrow$ eccentric contraction $\rightarrow$ weight plate $\rightarrow$ knurling $\rightarrow$ deep drawing $\rightarrow$ hydroforming $\rightarrow$ direct metal laser sintering $\rightarrow$ rapid prototyping $\rightarrow$ \textbf{3D printing}  &  \textbf{health} $\rightarrow$ health care $\rightarrow$ palliative care $\rightarrow$ intensive care unit $\rightarrow$ \textbf{3D printing}  \\ \bottomrule
    \end{tabular}

\end{table}


\begin{figure*}[t!]
	\centering
		\begin{tabular}{m{1cm}m{4.5cm}<{\centering}m{4.5cm}<{\centering}m{4.5cm}<{\centering}}
        ~ & Combined relationship & Statistical relationship & Semantic relationship\\
        General & {\includegraphics[width=4.5cm]{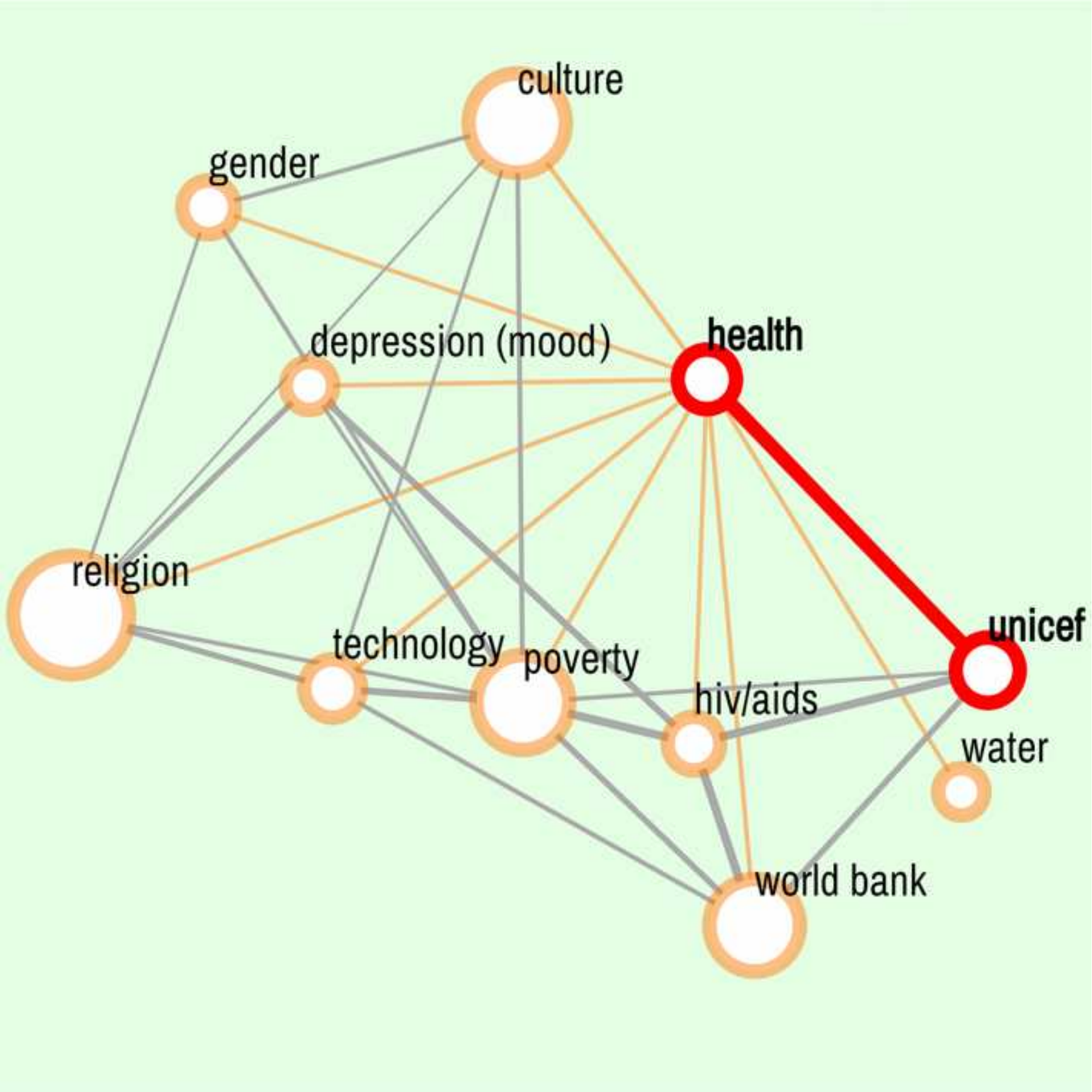}} & {\includegraphics[width=4.5cm]{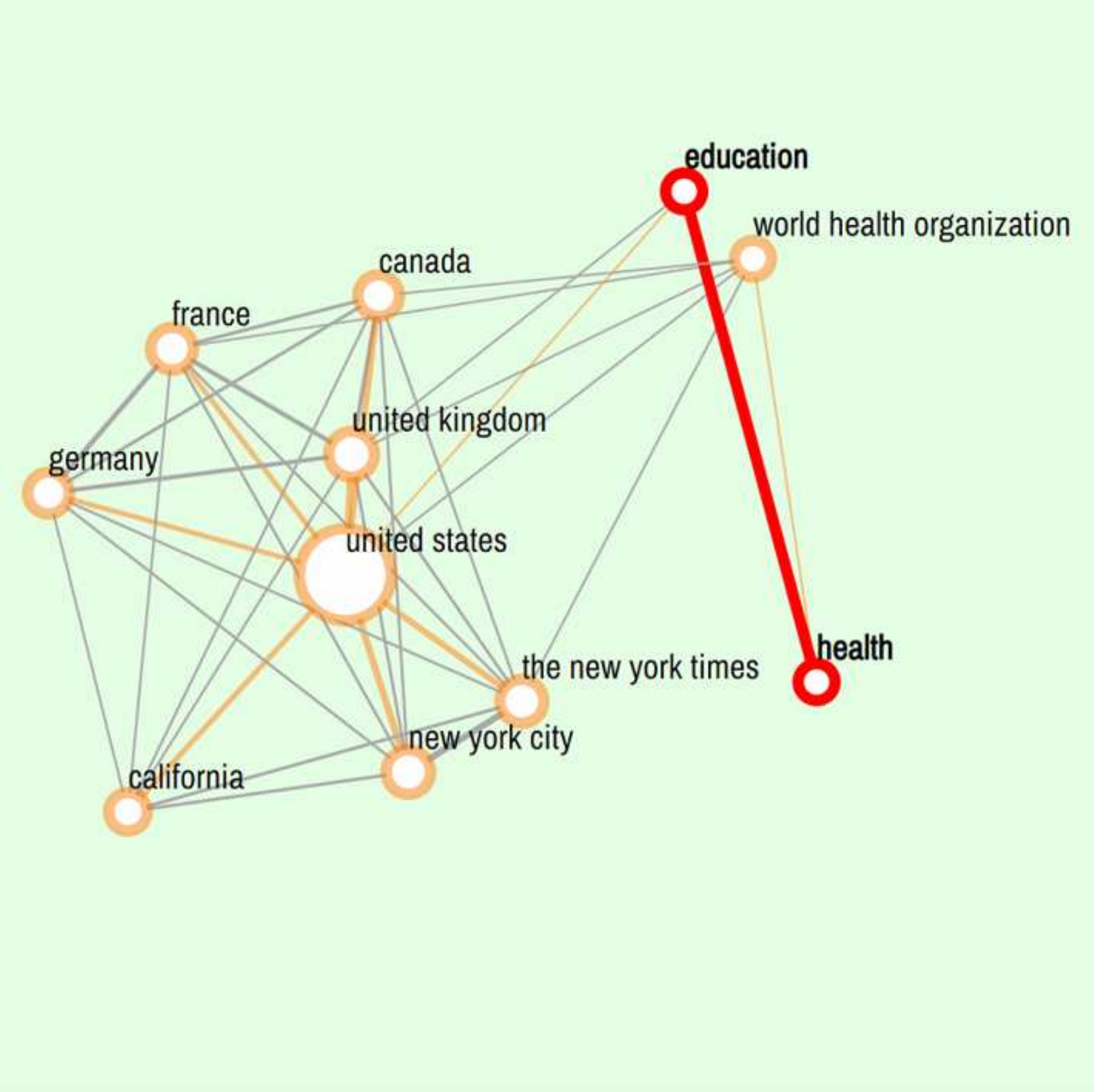}} & {\includegraphics[width=4.5cm]{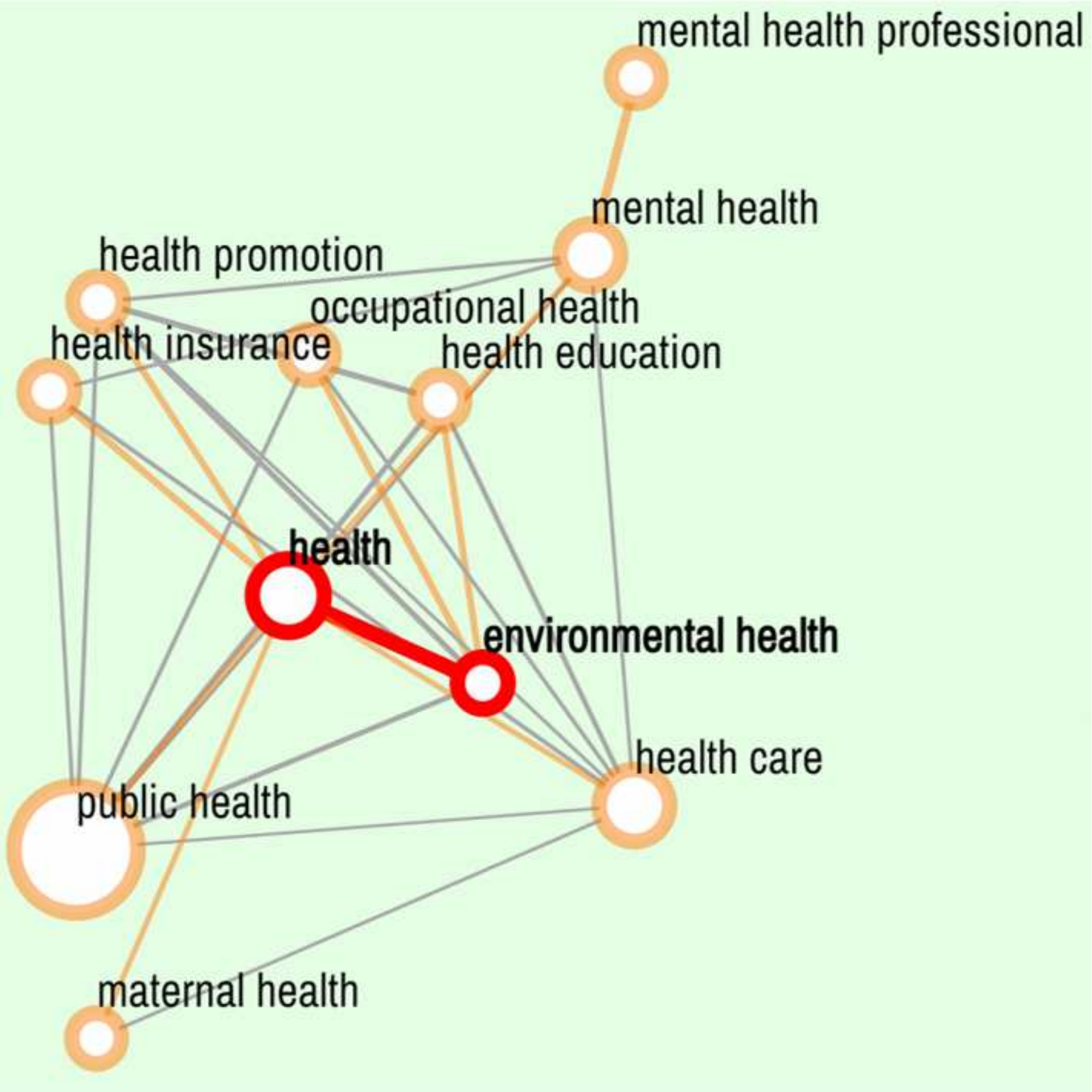}}\\
        Specific & {\includegraphics[width=4.5cm]{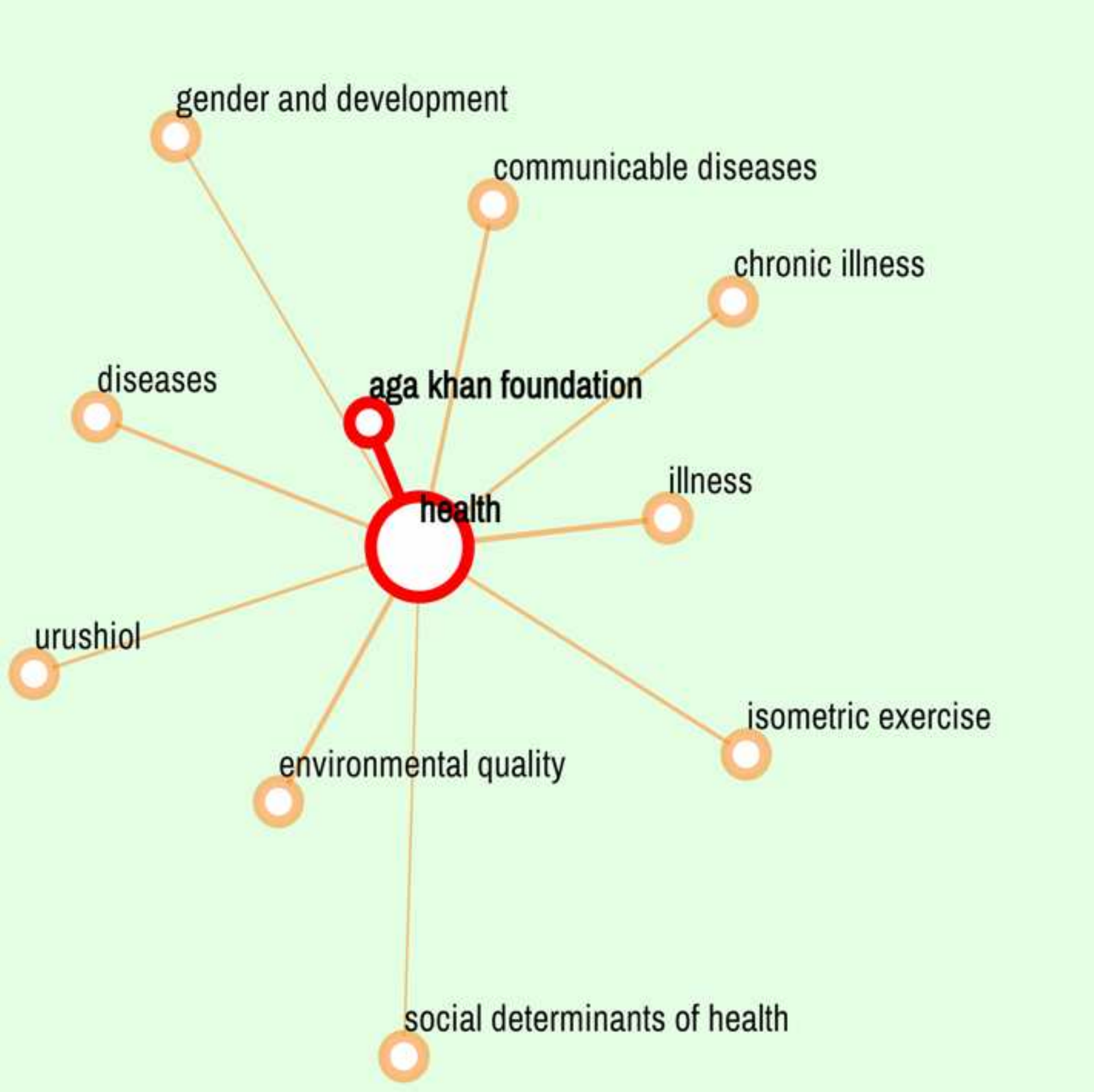}} & {\includegraphics[width=4.5cm]{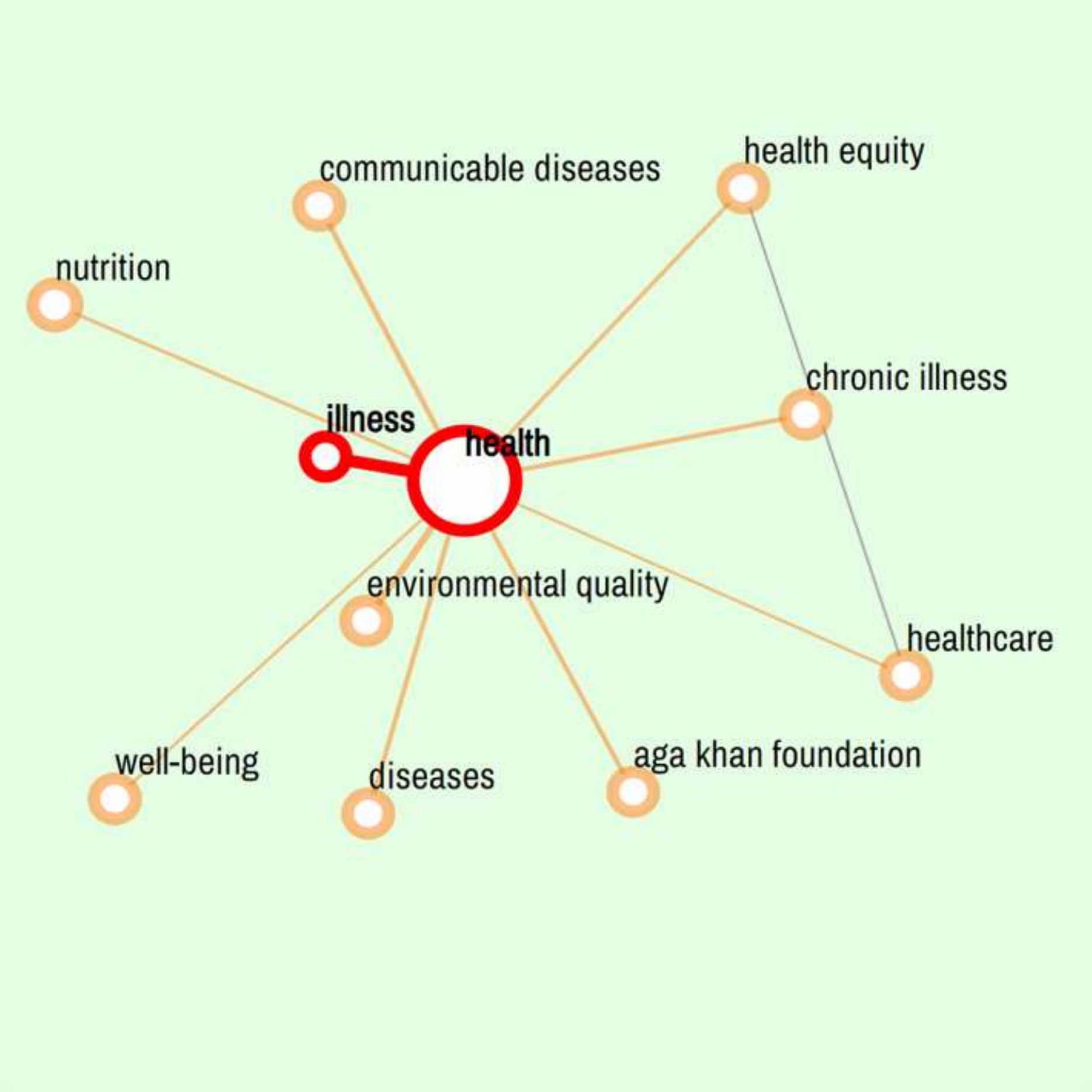}} & 
        {\includegraphics[width=4.5cm]{g_sem.pdf}}\\
        \end{tabular}
	\caption{Retrieval results for "health" with three different relationships}
	\label{FIG:weightCompareExplore}
\end{figure*}

\begin{table}[]
\caption{The average node degree of retrieval results for "health" with two different relationship}\label{Tlb:NDE}
\begin{tabular}{p{3.5cm}<{\centering}m{2cm}<{\centering}m{2cm}<{\centering}}
\toprule
Category & "Explore-General" & "Explore-Specific" \\ \midrule
    Statistical relationship & 536 & 63\\
       Combined relationship & \textbf{308} & \textbf{32}\\
       \bottomrule
    \end{tabular}
\end{table}

\begin{table}[t!]
\caption{The average node degree of knowledge associations between "health" and "3d printing" with two different relationship}\label{Tlb:NDS}
\begin{tabular}{p{3.5cm}<{\centering}m{2cm}<{\centering}m{2cm}<{\centering}}
\toprule
Category & "Search Path-Basic" & "Search Path-Professional" \\ \midrule
    Statistical relationship & 565 & 139\\
       Combined relationship & \textbf{473} & \textbf{131}\\
       \bottomrule
    \end{tabular}
\end{table}

Figure \ref{FIG:weightCompareExplore} and Table \ref{Tlb:KA} are the results of “Explore” and “Search Path” respectively. It can be seen that the results of “Explore” and “Search Path” with statistical relationship have more concepts which contain common and general meaning but are irrelevant with “health” semantically, e.g., “United States” and “United Kingdom” which are dominant nodes in this case. Conversely, the results of the two functions with semantic relationships contain more relevant concepts but only show the semantic relevance to “health” (e.g., “environmental health” and “health care”). The combined relationship makes a balance between the statistical relationship and semantic relationship so that it produces a relatively positive result. The node degree of a concept means the sum of weights of all edges incident to that node. The average node degree of concepts are calculated in combined relationships and statistical relationships to demonstrate whether the very common results are balanced quantitatively. Table \ref{Tlb:NDE} and Table \ref{Tlb:NDS} shows that, in four functions, the average node degrees of concepts with combined relationship are all observably lower than that of concepts with statistical relationship, which imply that the semantic relationship balances the statistical relationship to retrieve valuable information. Both the quantitative and qualitative results indicate that the combined relationship is efficient to reduce the influence of dominant concepts with high node degree in retrieval results thus could facilitate design innovation activities. 

\section{Demonstration}
In this section, we showcase four functions in WikiLink for information retrieval and design innovation. Qualitative analysis of the results is performed to demonstrate the features of each function. In addition, the performance of WikiLink is compared with four state-of-the-art tools, and the corresponding results are also analyzed qualitatively.
\begin{figure*}
	\centering
		\begin{tabular}{m{1cm}m{6.2cm}<{\centering}m{6.2cm}<{\centering}}
        ~ & "3d printing" & "fused deposition modeling"\\
        General & {\includegraphics[width=6cm]{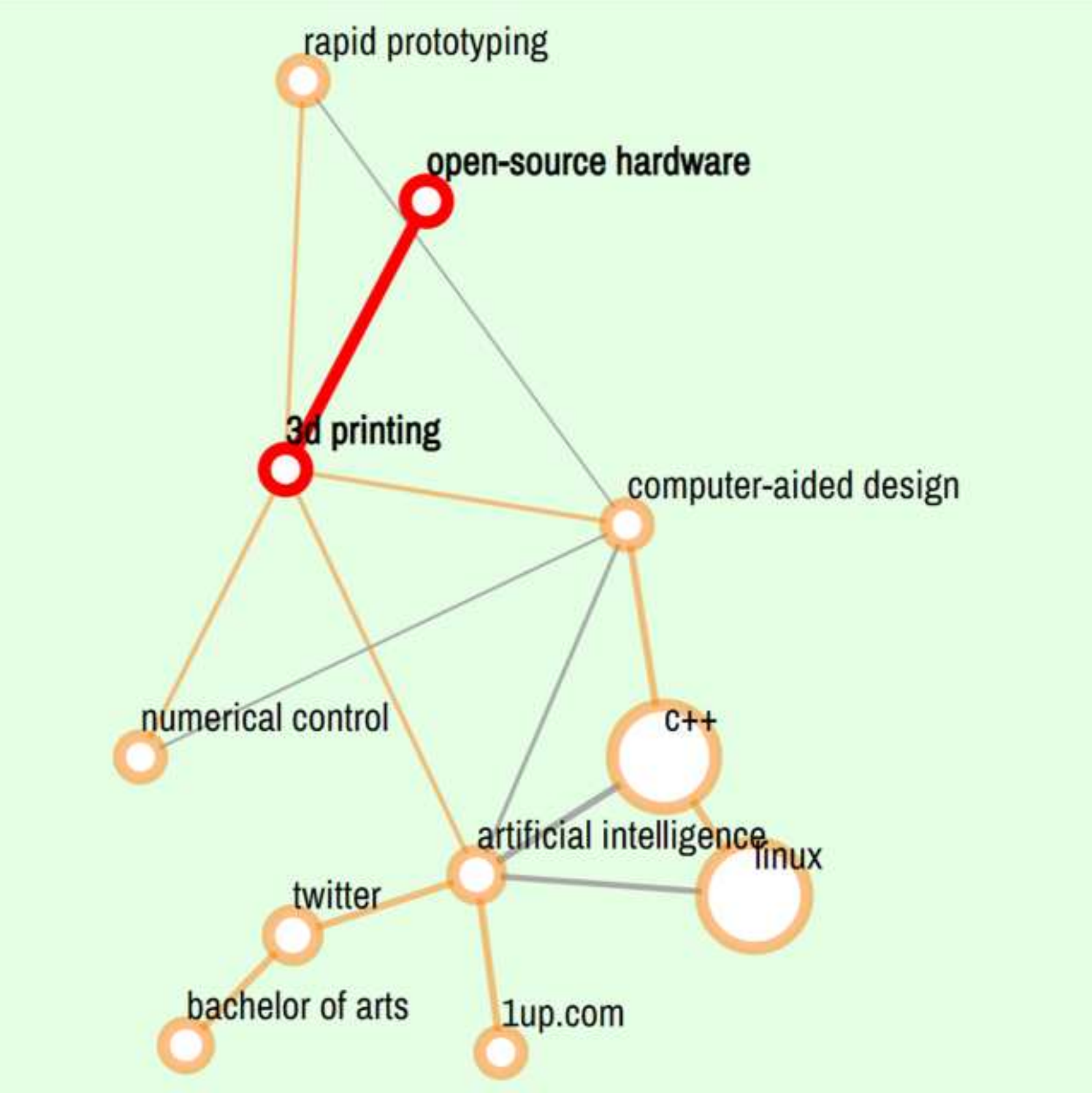}} & {\includegraphics[width=6cm]{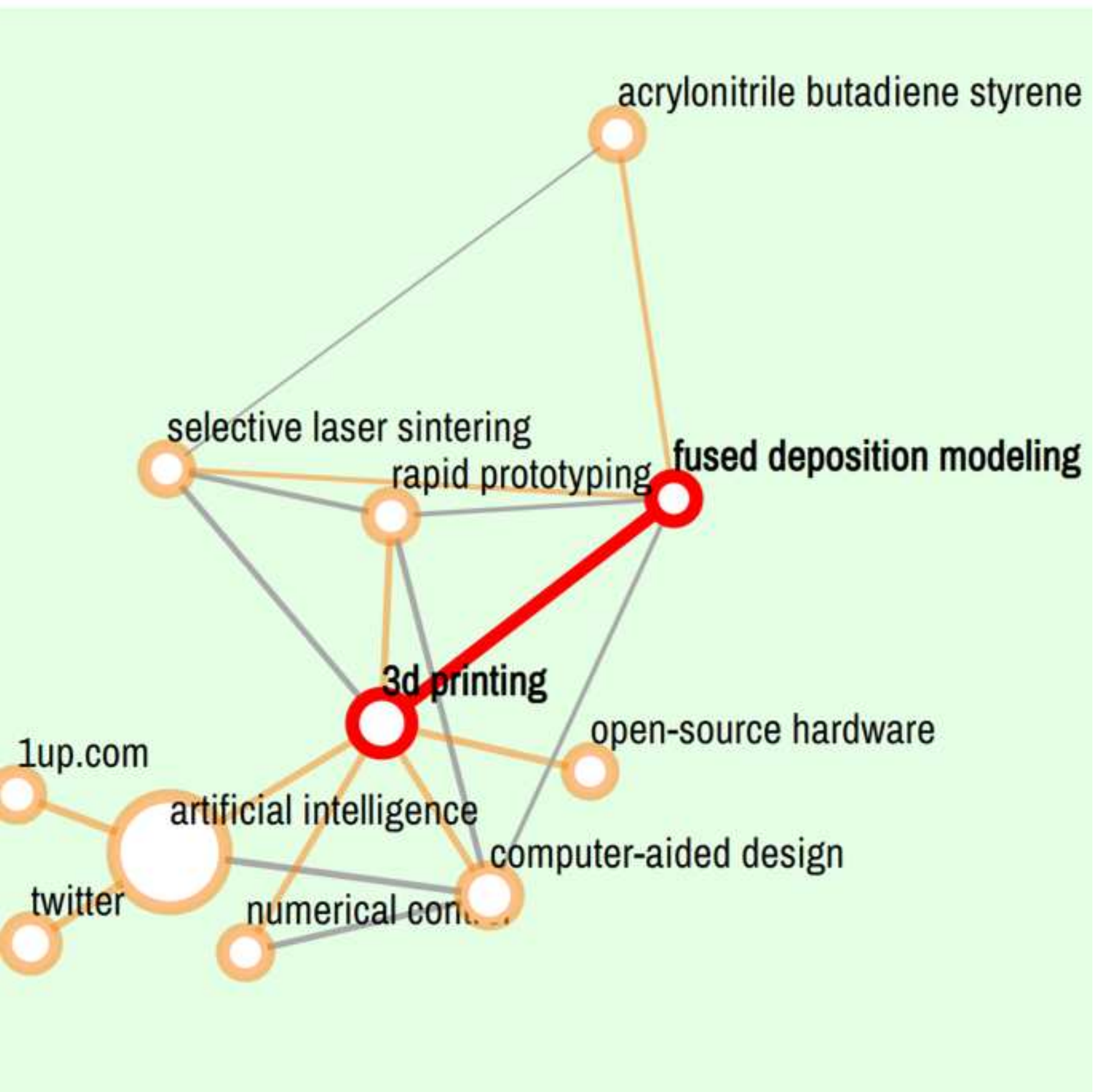}}\\
        Specific & {\includegraphics[width=6cm]{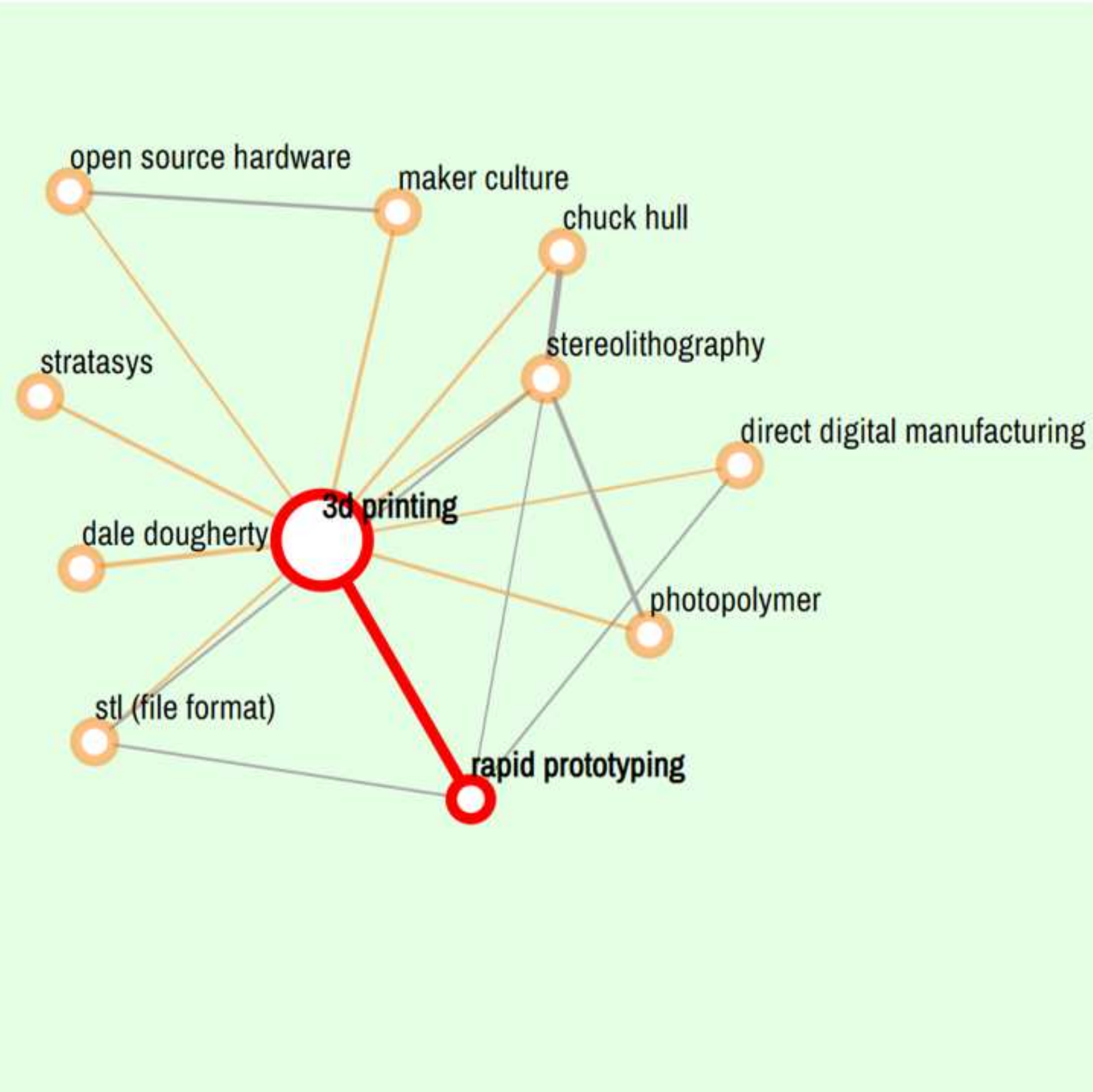}} & {\includegraphics[width=6cm]{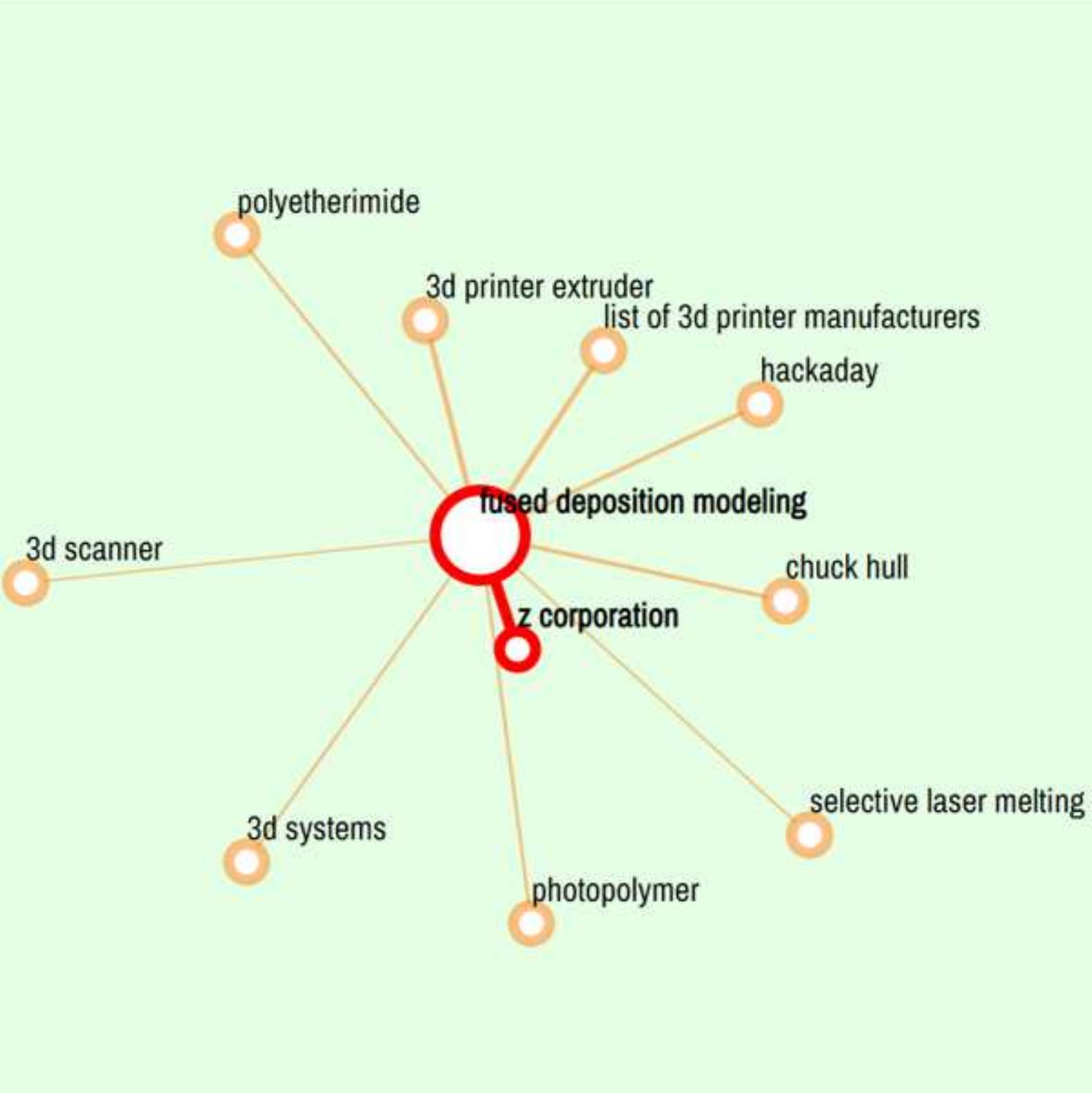}}\\
        \end{tabular}
	\caption{Retrieval results for "3d printing" and "fused deposition modeling" in "Explore-General" and "Explore-Specific" mode}
	\label{FIG:explore}
\end{figure*}
\subsection{The "Explore-General" and "Explore-Specific" mode }
To fairly compare the performance of "Explore-General" and "Explore-Specific" modes, two terms in the field of engineering design are chosen: "3d printing" and "fused deposition modeling". 3D printing is a multi-faceted technology and has been employed across a broad range of applications \citep{berman20123}, and is a widely used term with general meanings. Fused deposition modeling (FDM) is a 3D printing method that heats a continuous thermoplastic filament and extrudes it for layer-by-layer deposition \citep{hamzah20183d}, which is also a widely used term with specific meanings. These two terms are inputted and explored in "general" and "specific" modes, respectively. Figure \ref{FIG:explore} shows the top 10 relevant terms in each retrieval. By comparing the "general" results (the first row) with the "specific" results (the second row), it can be seen that the terms in "general" results are more common and comprehensible, such as computer-aided design, and artificial intelligence, while the terms in "specific" results, such as stl (file format) and polyetherimide, are normally very specific concepts in particular domains. Furthermore, as the figure shows, FDM’s specific result is centered scattering. It implies that primary terms in a particular domain are discrete and irrelevant to each other.

\begin{table}
\caption{The high-correlated two types of knowledge associations}
\label{tab:association}
\begin{tabular}{p{1.5cm}<{}p{6.5cm}p{6.5cm}}
\toprule
~ & Brain \& Computer & Avocado \& Chair\\ \midrule
       \multirow{3}{*}{Basic} & \textbf{brain} $\rightarrow$ artificial intelligence $\rightarrow$ \textbf{computer} & \textbf{avocado} $\rightarrow$ fruit $\rightarrow$ furniture $\rightarrow$ \textbf{chair} \\
        ~ & ~ & ~ \\
        ~ & \textbf{brain} $\rightarrow$ biology $\rightarrow$ \textbf{computer} & \textbf{avocado} $\rightarrow$ walnut $\rightarrow$ furniture $\rightarrow$ \textbf{chair} \\ \hline
        \multirow{5}{*}{Professional} & \textbf{brain} $\rightarrow$ neuroscience $\rightarrow$ psychology $\rightarrow$ science $\rightarrow$ technology $\rightarrow$ \textbf{computer} & \textbf{avocado} $\rightarrow$ guacamole $\rightarrow$ burrito $\rightarrow$ xylitol $\rightarrow$ product call $\rightarrow$ ikea $\rightarrow$ rocking chair $\rightarrow$ \textbf{chair} \\
        ~ & ~ & ~ \\
        ~ & \textbf{brain} $\rightarrow$ neuroscience $\rightarrow$ psychology $\rightarrow$ science $\rightarrow$ technology $\rightarrow$ internet  $\rightarrow$ \textbf{computer} & \textbf{avocado} $\rightarrow$ guacamole $\rightarrow$ taco $\rightarrow$ hockey puck $\rightarrow$ potato chips $\rightarrow$ ladder $\rightarrow$ rocking chair  $\rightarrow$ \textbf{chair} \\ \bottomrule
    \end{tabular}
\end{table}

\subsection{The "Search Path-Basic" and "Search Path-Professional" mode}
The "Search Path" function allows users to explore the implicit associations between two items even from different domains. It also has two modes that can return two types of associations. In order to test the above two modes, we used two pairs of terms, "brain" and "computer", which are weakly related, and "avocado" and "chair", which are seemingly unrelated. Table \ref{tab:association} shows the retrieved highest-correlated "basic" and "professional" knowledge associations of the two pairs. Obviously, the "basic" paths are shorter and the "professional" paths are longer. Besides, most of the nodes in "basic" paths are concepts with general meanings between the two domains, such as artificial intelligence, fruit and furniture, while the "professional" path is longer and the nodes are almost scientific terms or specific objects such as "neuroscience", "xylitol" and "guacamole". Some explicit associations are discovered in the results. For example, brain science drives the advance of computer science, especially artificial intelligence, which appears in the path "brain → artificial intelligence → computer". In addition, more implicit associations are connected by some surprising concepts like "fruit", "furniture" and "rocking chair", which may remind the idea of fruit-shaped furniture such as an avocado-shaped rocking chair. It is found that, in some cases, purely statistical weights between edges result in a longer and more surprising path which may inspire more innovative ideas in design activities.

\begin{figure*}[t!]
		\begin{tabular}{p{5.3cm}p{5.3cm}p{5.3cm}}
        {\includegraphics[width=5.3cm]{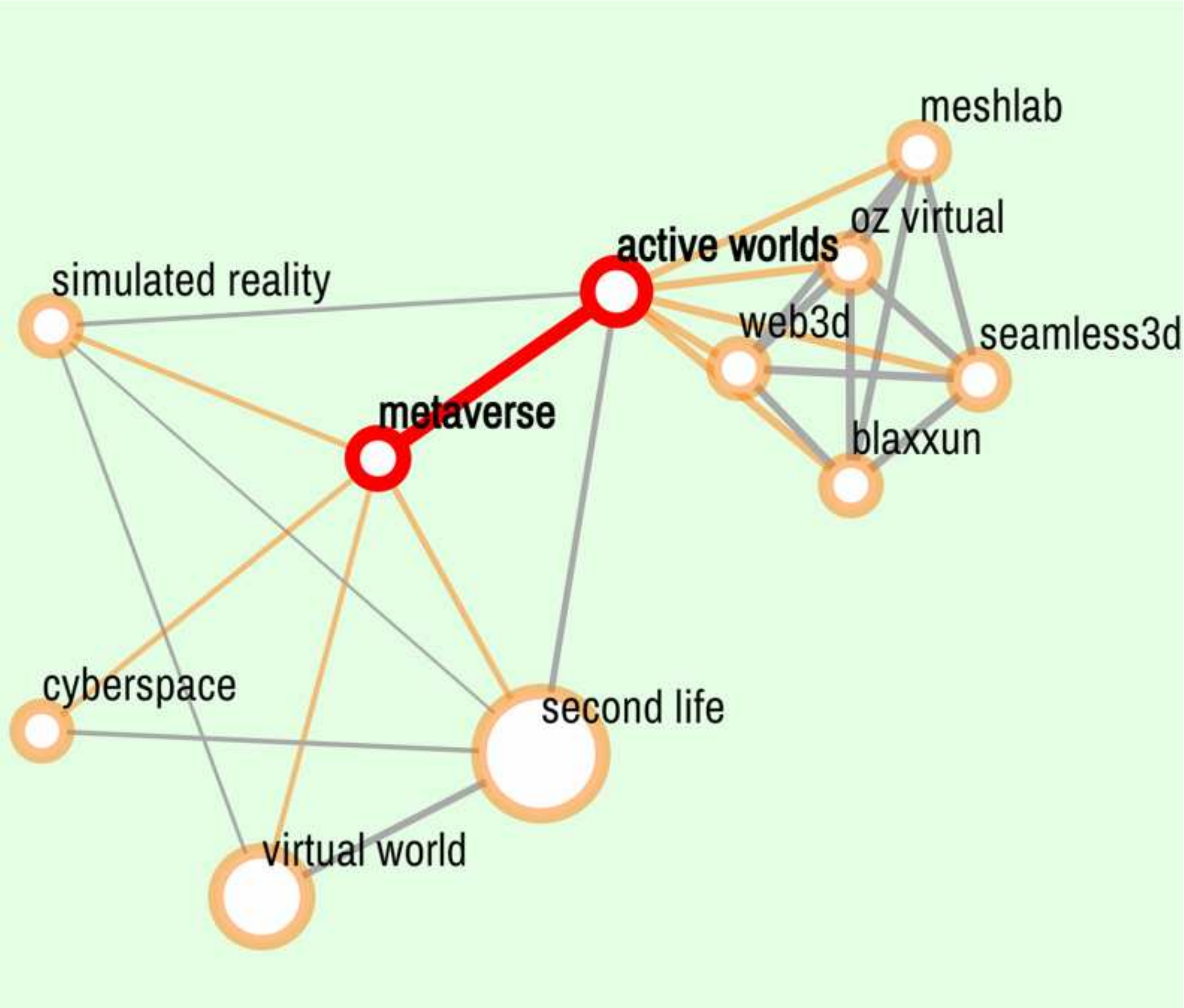}} & {\includegraphics[width=5.3cm]{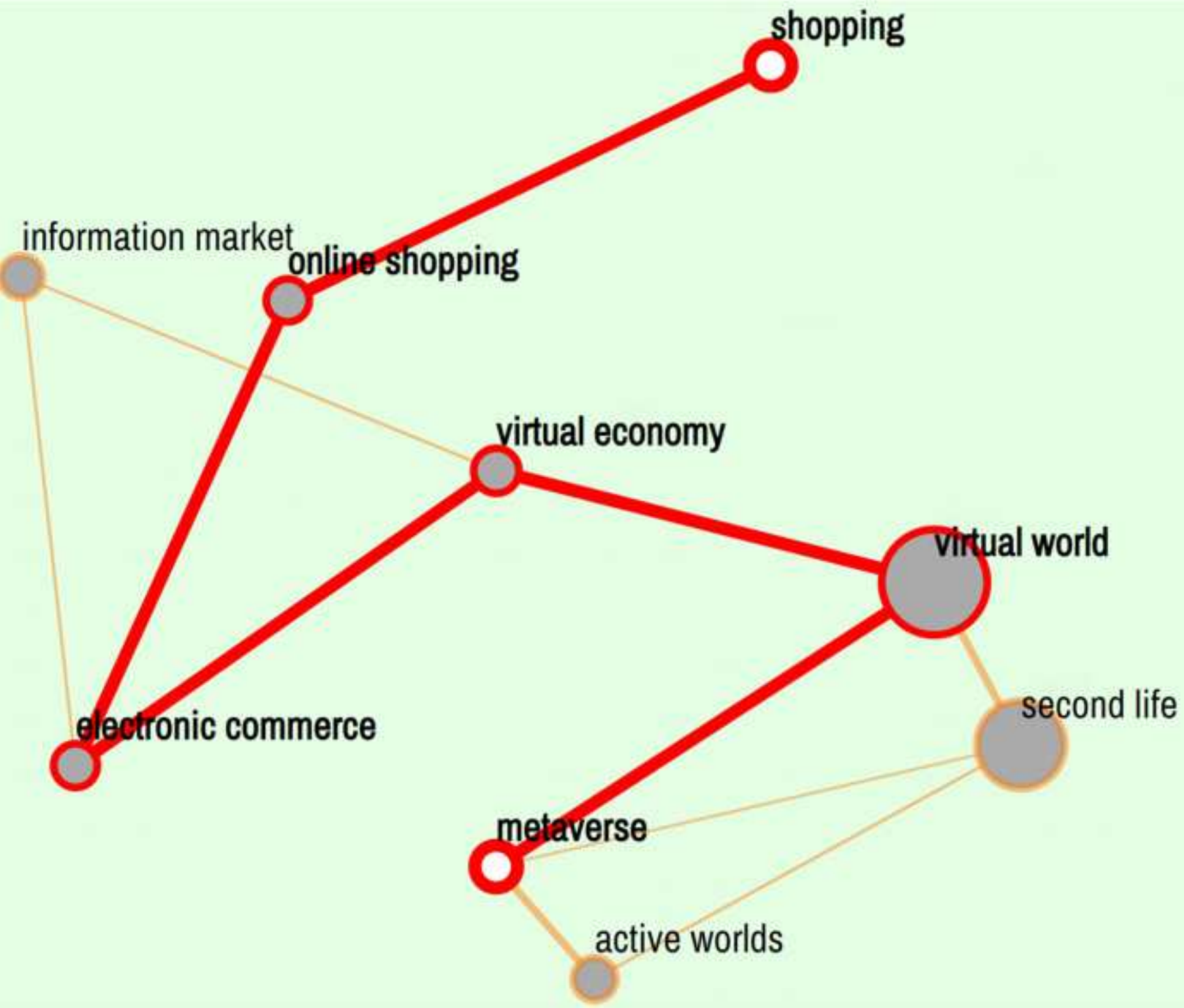}} & {\includegraphics[width=5.3cm]{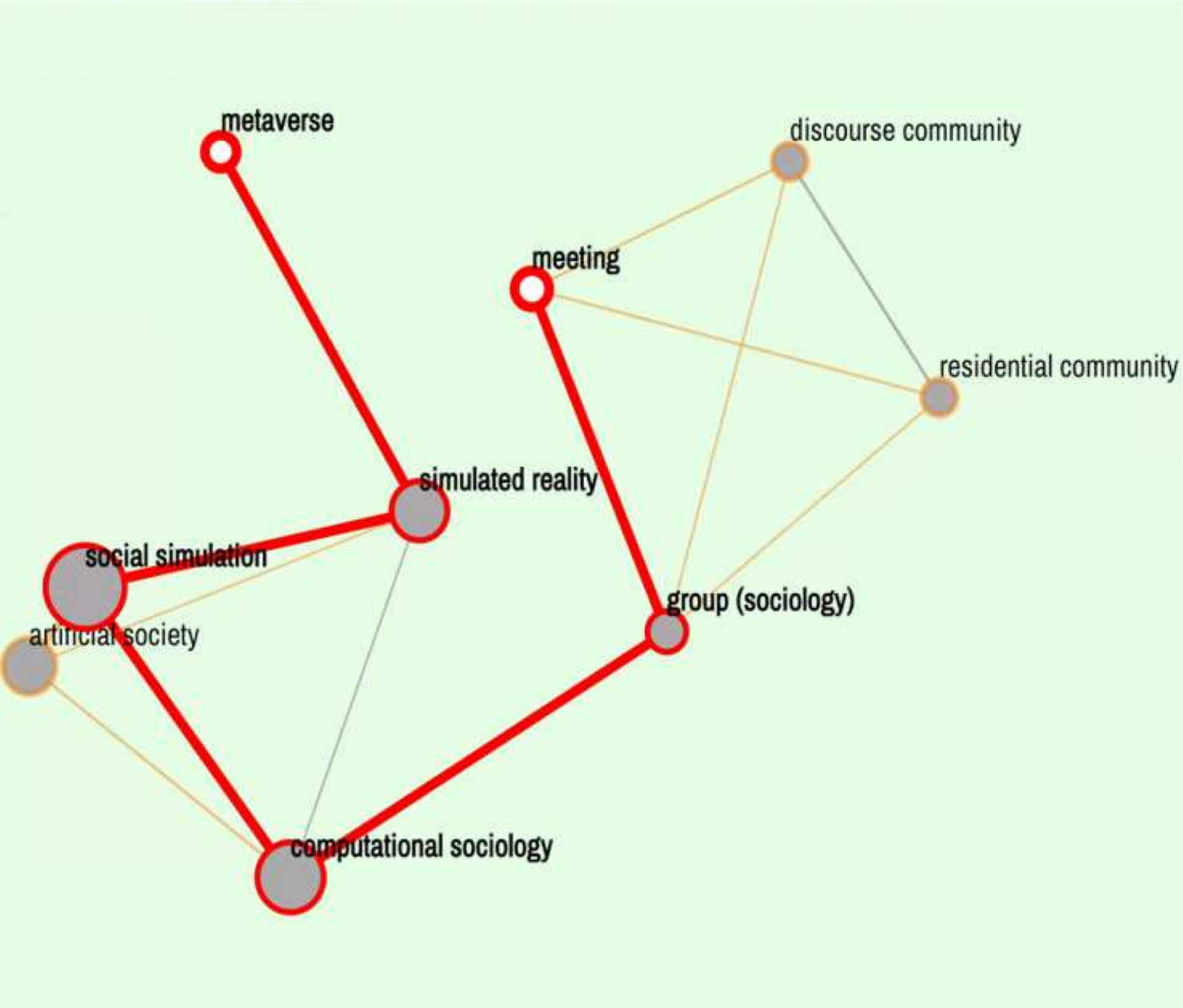}}\\
       (a) The "Explore Specific" results for "metaverse" & (b) The "Search Path-Professional" results between "metaverse" and "shopping" & (c) The "Search Path-Professional" results between "metaverse"  and "meeting"\\        \end{tabular}
	\caption{Comparisons of the results from "Explore" and "Search path"}
	\label{FIG:explo and find1}
\end{figure*}

\subsection{The "Explore" and "Search Path" function}
The above shows that the "Explore" function aims to discover the knowledge associations around a single term, while the "Search Path" function aims to search for the associations between two terms. To clarify the difference between them, a hot concept in engineering design, "metaverse", was explored along with two weakly related terms separately: "shopping" and "meeting". Retrieval experiments were conducted in "Explore-Specific" and "Search Path-Professional" mode respectively. As shown in Figure \ref{FIG:explo and find1}, the retrieval results of "metaverse" cover a wide range of fields including "virtual world", "simulated reality", "cyberspace" and related games including "Second Life" and "Active Worlds". These wide results can lead to comprehensive knowledge discovery and an open imagination about the target term. On the other hand, the paths between "metaverse" and a selected concept focus on bridging the fields that connect them, which leads to combinational ideas. For instance, the nodes linking "metaverse" and "shopping" are related to "virtual economy", and the nodes linked "metaverse" and "meeting" are related to virtual society.

\begin{table}[]
\caption{The top 10 related terms to "neural network" and "trypsin" in WikiLink and the four benchmark tools}
\label{tab:benchmark1}
\begin{tabular}{m{1.5cm}<{\centering}m{7cm}m{7cm}}
\toprule
~ & Neural Network & Trypsin \\ \midrule
       WikiLink (general) & deep learning, google, c++, linux, cross-platform, javascript, open-source software, operating system, perl & amino acid, pancreas, enzyme, transcription(genetics), translation(genetics), base pair(genetics), life, active site, translation(biology), stroke \\\hline
       WikiLink (specific) & classification rule, deep learning, cognitive model, stockfish(chess), machine learning, black box, Hebbian learning, list of memory biases, deepmind, artificial neuron & phenylisothiocyanate, myotoxin, triosephosphateisomerase, zymogen, tandem mass spectrometry, peptide mass fingerprinting, ligase, dihydrofolate reductase, pepsin, papain \\\hline 
        B-link (general) & genetic algorithm, optimization, fuzzy logic, classification, pattern recognition, artificial neural network, multi-objective optimization, simulated annealing, simulation, response surface methodology & chymotrypsin, protease, pepsin, purification, thrombin, digestion, characterization, expression, synthesis, crystal structure \\\hline
        B-link (specific) & backpropagation, genetic algorithm, fuzzy logic, self-organizing map, multilayer perceptron, backpropagation algorithm, neuro-fuzzy, pattern recognition, neuro-fuzzy system, artificial intelligence & chymotrypsin, enzyme thermostability, modified enzyme, pepsin, protease-activated receptor-2, protease-activated receptor, digestive protease, pyloric caecum, carboxypeptidase a, viscera \\\hline
       ConceptNet & neural net; autoencoder, backpropagation, catastrophic interference, computational intelligence, condela, convolutional neural network, dropout, hidden layer &	antitrypsin, antitryptic, apronitin, chymotrypsin, endopeptidase, enterokinase, meromyosin, mesotrypsin, ovoinhibitor, ovomucin \\\hline
        WordNet & neural net, computer architecture, network of neurons, network of nuclei & enzyme, pancreas, protein, polypeptide units\\\hline
        TechNet & artificial neural network, machine learning, training data, pattern recognition,  hidden layer, layer node, upper hidden layer, neuron, residual activation, automobile overspeed, vehicular safety sensor, time many & 	Proteolytic enzyme, pepsin trypsin, subtilisin family, bromelain ficin, proteolytic, no amidolytic, enzymatic, amidolytic, protease, trypsin thrombin plasmin, dynorphin targeting moiety, irtx 
        \\ \bottomrule
    \end{tabular}
\end{table}
\subsection{Comparison with benchmark tools}
We undertook a retrieval comparison between WikiLink and the other four benchmark tools. The target terms are "neural network" in computer science and "trypsin" in medical physiology. The aim of this experiment is to test whether our network can return a broad range of related terms which are able to stimulate innovation in the design process efficiently. Since the number and presentation of retrieval results vary from tool to tool, we have selected the top 10 related terms for each tool to present in Table \ref{tab:benchmark1}. Especially, the results of WikiLink and B-link \citep{shi2017data} were obtained through their "Explore" function, the result of ConceptNet \citep{speer2017conceptnet} was obtained from its "Related terms" category, the result of WordNet \citep{fellbaum2010wordnet} was obtained from its "Synset" and "Example sentence" functions. According to Table \ref{tab:benchmark1}, the terms retrieved by WikiLink in the "general" and "specific" modes both prove the effectiveness of the "Explore" function. For example, the retrieval results of "neural network" in the "specific" mode are all domain-specific terms related to the components (e.g., "artificial neuron"), functions (e.g., "cognitive model") and applications (e.g., "deep learning") of "neural network". Since the "Explore" function of WikiLink is divided into the "Explore-General" and "Explore-Specific" modes, its results, containing common terms (from the "Explore-General" mode) and technical terms (from the "Explore-Specific" mode), cover a comprehensive range. In contrast, ConceptNet, WordNet and TechNet simply have only one retrieval mode, which leads to their retrieved results invariably focus on some technical terms in a specific range. Even though B-Link retrieves in the two modes as WikiLink, its results are also limited by the data source which are engineering academic papers and design websites. It can be seen that the retrieved terms of B-Link tend to contain specific meanings. Instead, WikiLink applies Wikipedia as the data source for its semantic network, which covers information from a wide range of domains.
The comparison suggests that WikiLink is more capable of retrieving terms in various domains, which is essential for knowledge discovery in the knowledge-intensive design innovation process.

\subsection{A design case}
\begin{figure*}[t!]
	\centering
	\includegraphics[width=0.9\textwidth]{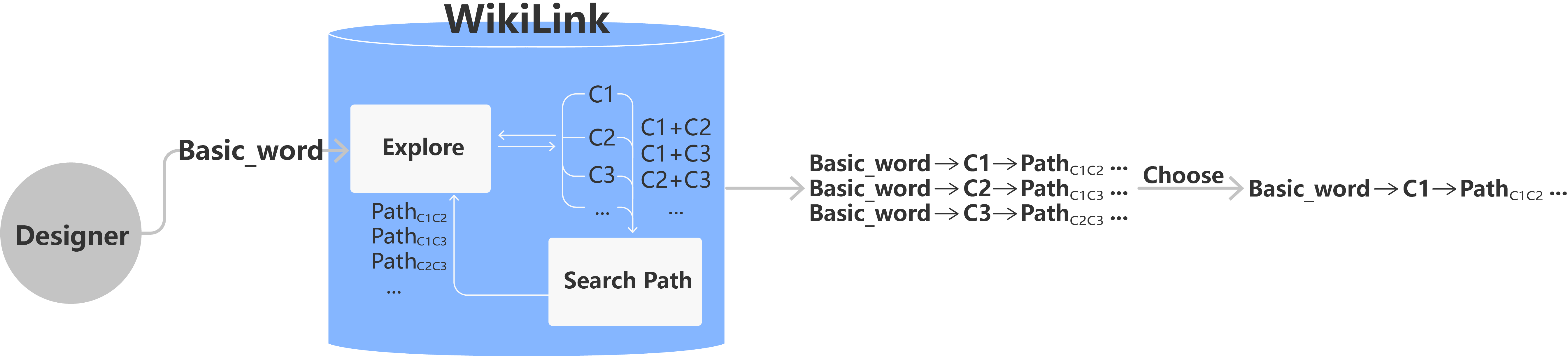}
	\caption{The flow of concepts exploration in WikiLink} 
	\label{FIG:flow}
\end{figure*}

A designer is recruited to conduct a design case and demonstrate the process of applying WikiLink for design innovation. Generally, the designer would be initially given a design question with a "Basic word", and then required to apply the "Explore" function and "Search path" function in WikiLink to freely explore the related concepts around the "Basic word" which could potentially inspire the designer. By applying the "Explore function", the designer could discover the knowledge concepts "C1", "C2", "C3" around the "Basic word" as denoted in Figure \ref{FIG:flow}. While the "Search path" function provides the paths, e.g \textit{Path}\textsubscript{\textit{C1C2}} between two terms "C1" and "C2" for combinational creativity \citep{han2019further}. This process can be iteratively applied to discover knowledge associations and paths such as "C3" and \textit{Path}\textsubscript{\textit{C1C3}}. The related concepts obtained from WikiLink are then used to form design inspiration links such as "Basic\_word-C1, \textit{Path}\textsubscript{\textit{C1C3}},  \textit{Path}\textsubscript{\textit{C2C3}}" and some of them are eventually chosen for the design output of specific design ideas. 


\begin{figure*}[t!]
		\begin{tabular}{p{5.3cm}p{5.3cm}p{5.3cm}}

        {\includegraphics[width=5.3cm]{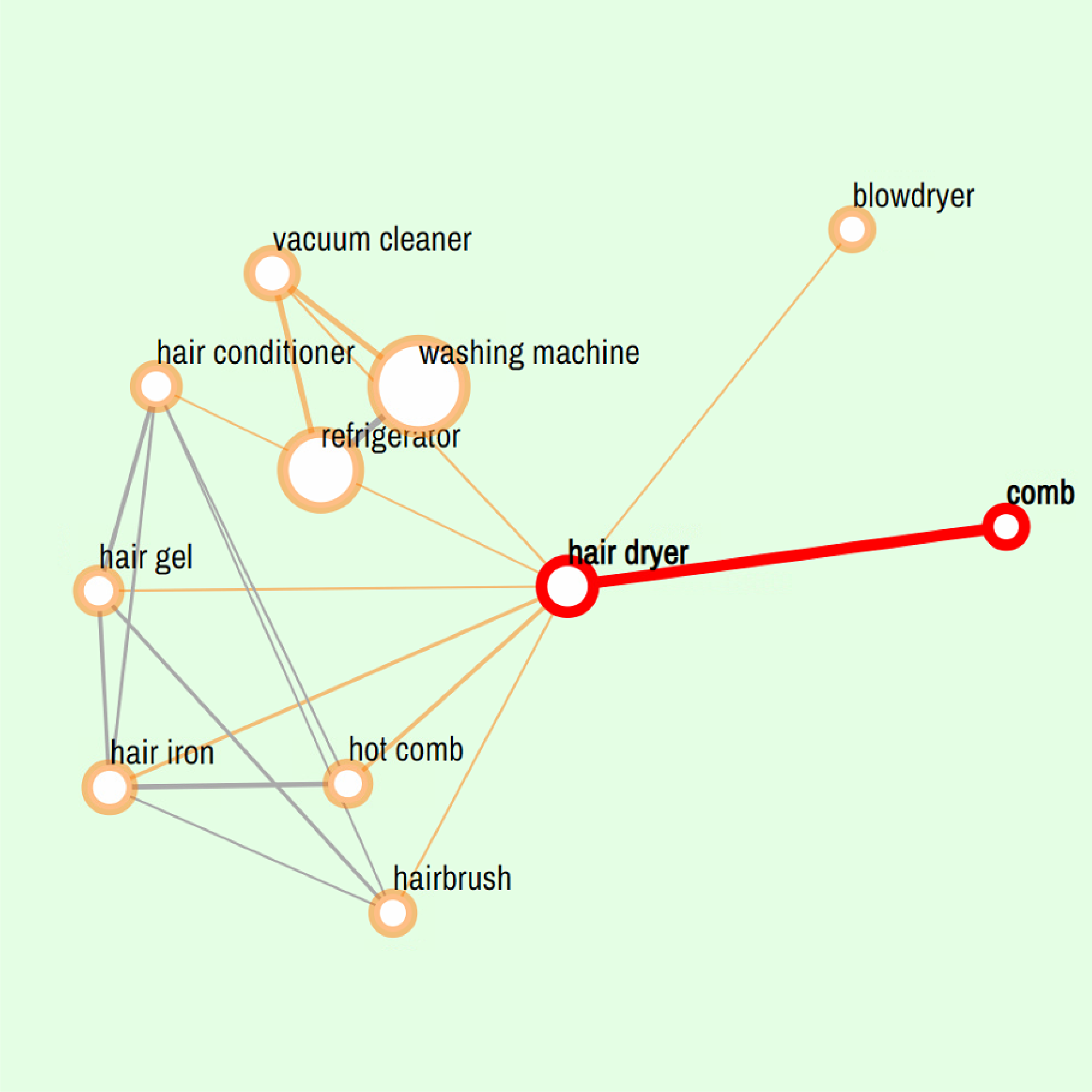}} & {\includegraphics[width=5.3cm]{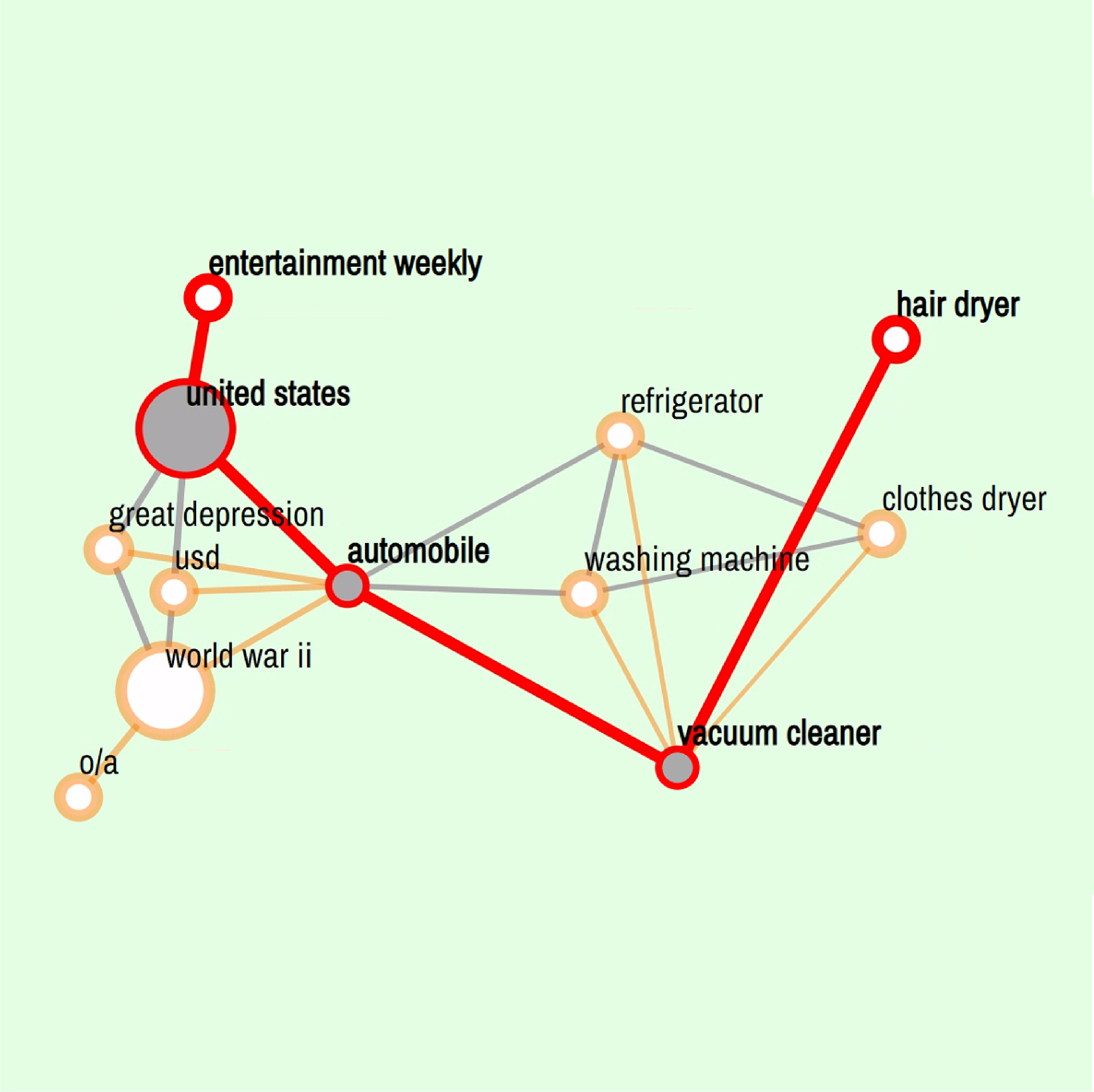}} & {\includegraphics[width=5.3cm]{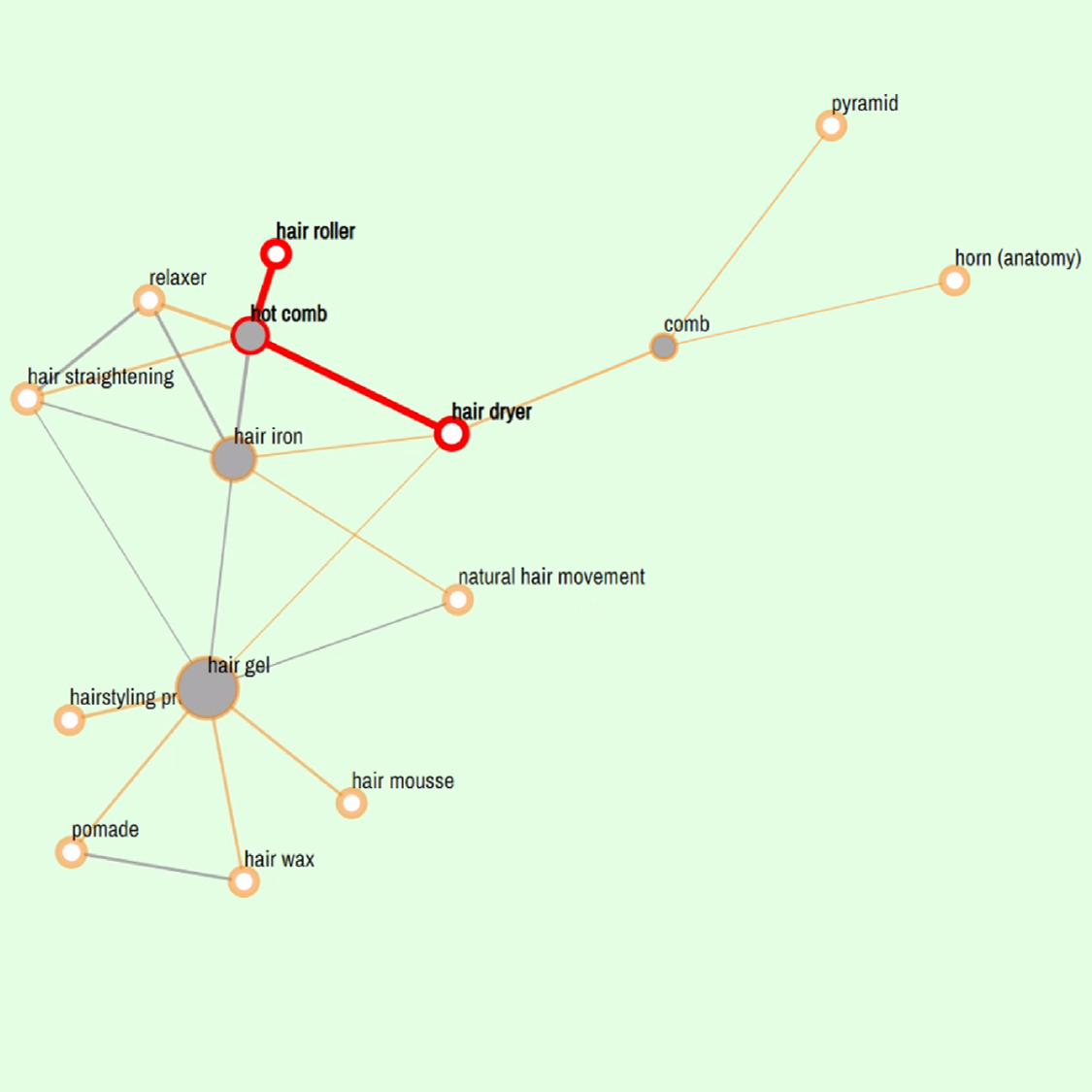}}\\
       (a) "Explore-General" results with one step for "hair dryer" & (b) "Explore-General" results with two steps for "hair dryer" & (c) "Explore-Specific" results with two step for "hair dryer"\\        \end{tabular}
	\caption{The examples of concepts retrieved by "Explore"}
	\label{FIG:explo and find}
\end{figure*}

A real design case is conducted to illustrate how to facilitate design innovation with WikiLink. Since "hair dryer" is a well-known product of which the homogenization in the market is serious and its innovation has encountered a bottleneck, the designer is required to generate ideas and provide innovative designs for hair dryer. The concept "hair dryer" is chosen as the design query (also known as the basic word) in WikiLink in this case. The designer then started with the "Explore" function by freely choosing several different step lengths and switching between general and specific mode for divergent and convergent thinking. Some screenshot examples are shown in Figure \ref{FIG:explo and find}. It is noted that the designer is not restricted to using "hair dryer" as the query only. 
After the initial exploration in WikiLink, the designer obtained some interesting and inspiring concepts, such as "Entertainment weekly", "Vacuum cleaner", "Comb","Hair iron"," Hair gel"," Hair roller"," Hot comb"," Horn" and "Pyramid". The next step is to apply the "Search Path" function by freely querying the paths between two concepts of the designer's interests. Some retrieval results are shown in Table \ref{tab:searchpathresult}.


\begin{table}[]
\caption{The paths between the inspiring concepts and "hair dryer" retrieved by "Search Path"}
\label{tab:searchpathresult}
\begin{tabular}{p{4.7cm}p{1.5cm}p{9cm}}
\toprule
\textbf{Query concepts} & \textbf{Mode} & \textbf{Retrieval results} \\ \midrule
hair dryer \& entertainment weekly & Basic & \textbf{hair dryer} $\rightarrow$ vacuum cleaner $\rightarrow$ automobile $\rightarrow$ united states $\rightarrow$ \textbf{entertainment weekly}                \\

hair dryer \& entertainment weekly&Professional&\textbf{hair dryer} $\rightarrow$ hair iron $\rightarrow$ natural hair  movement $\rightarrow$ afro $\rightarrow$ \textbf{tie-dye} $\rightarrow$ \textbf{zardozi} $\rightarrow$ choli $\rightarrow$ crop top $\rightarrow$ the face (magazine) $\rightarrow$ arena (magazine) $\rightarrow$ loaded (magazine) $\rightarrow$ fhm's 100 sexiest women (uk) $\rightarrow$ fhm $\rightarrow$ maxim (magazine) $\rightarrow$ people (magazine) $\rightarrow$ \textbf{entertainment weekly}                                                                                                              
\\
hair dryer \& tie-dye&Basic& \textbf{hair dryer} $\rightarrow$ vacuum cleaner $\rightarrow$ automobile $\rightarrow$ textile $\rightarrow$ \textbf{tie-dye} \\ \bottomrule
\end{tabular}
\end{table}

\begin{figure*}[t!]
	\centering
	\includegraphics[width=0.85\textwidth]{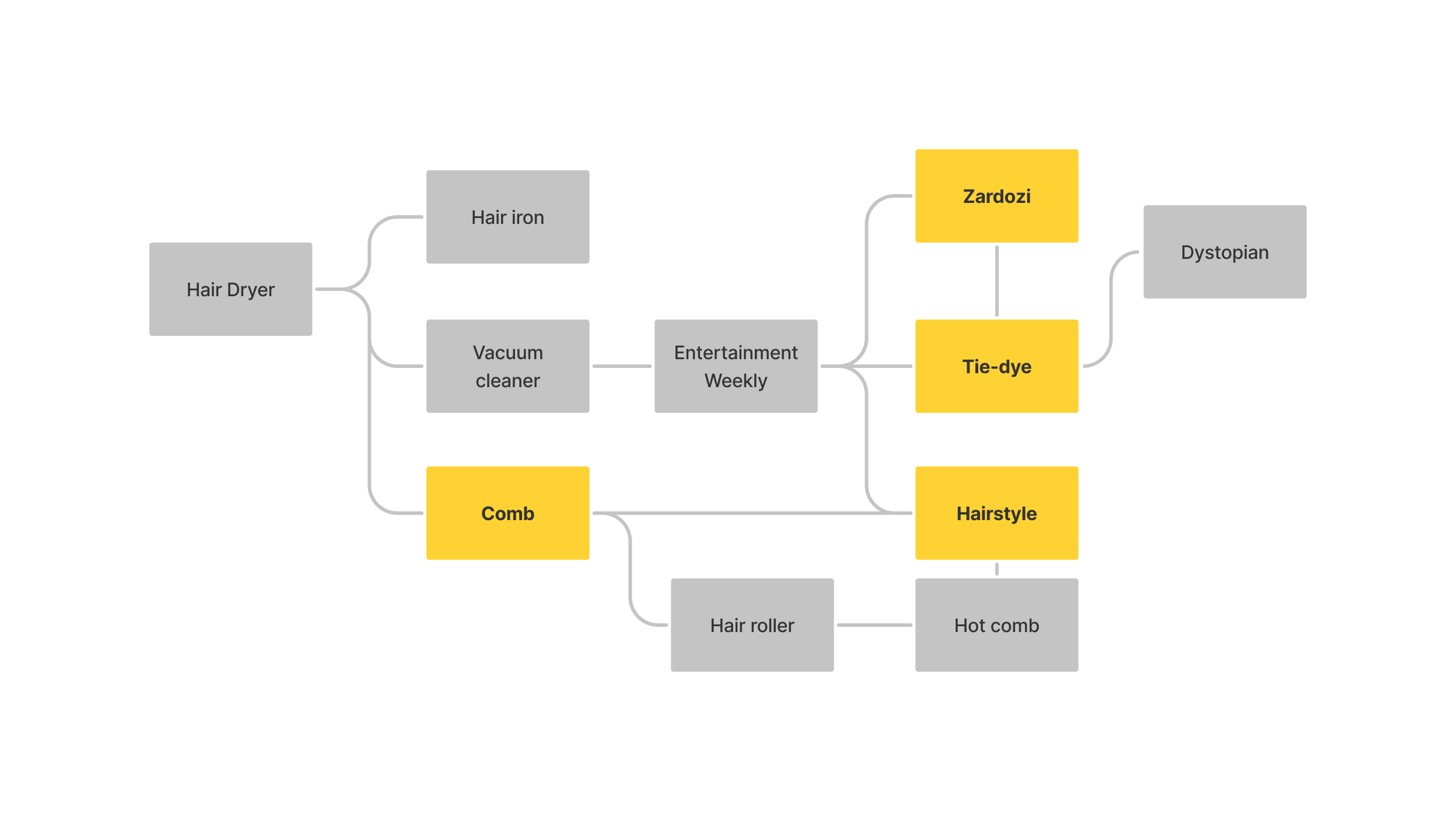}
	\caption{The example of “design inspiration link”} 
	\label{fig:inpirationlinks}
\end{figure*}

Afterwards, the designer continued to explore knowledge concepts for design innovation stimuli by iteratively using the "Explore" and "Search path" functions. The "Explore" function helps discover the knowledge associations around a single term, while the "Search path" function can potentially look for the associations between two terms. The designer recorded all the interesting and inspiring concepts and formed the "design inspiration links", as shown in Fig \ref{fig:inpirationlinks}, where the base of the link is "Hair Dryer", and rest of the concepts were from WikiLink obtained by using "Explore" and "Search path" functions. The above process was repeated to produce at least one design inspiration link and until the designer thought it is enough to formulate design ideas. Eventually, with the ideas originating from the concepts in the inspiration link, the designer produced the final complete design scheme and drew corresponding design sketches. 
 
\begin{figure*}[t!]
	\centering
	\includegraphics[width=0.4\textwidth]{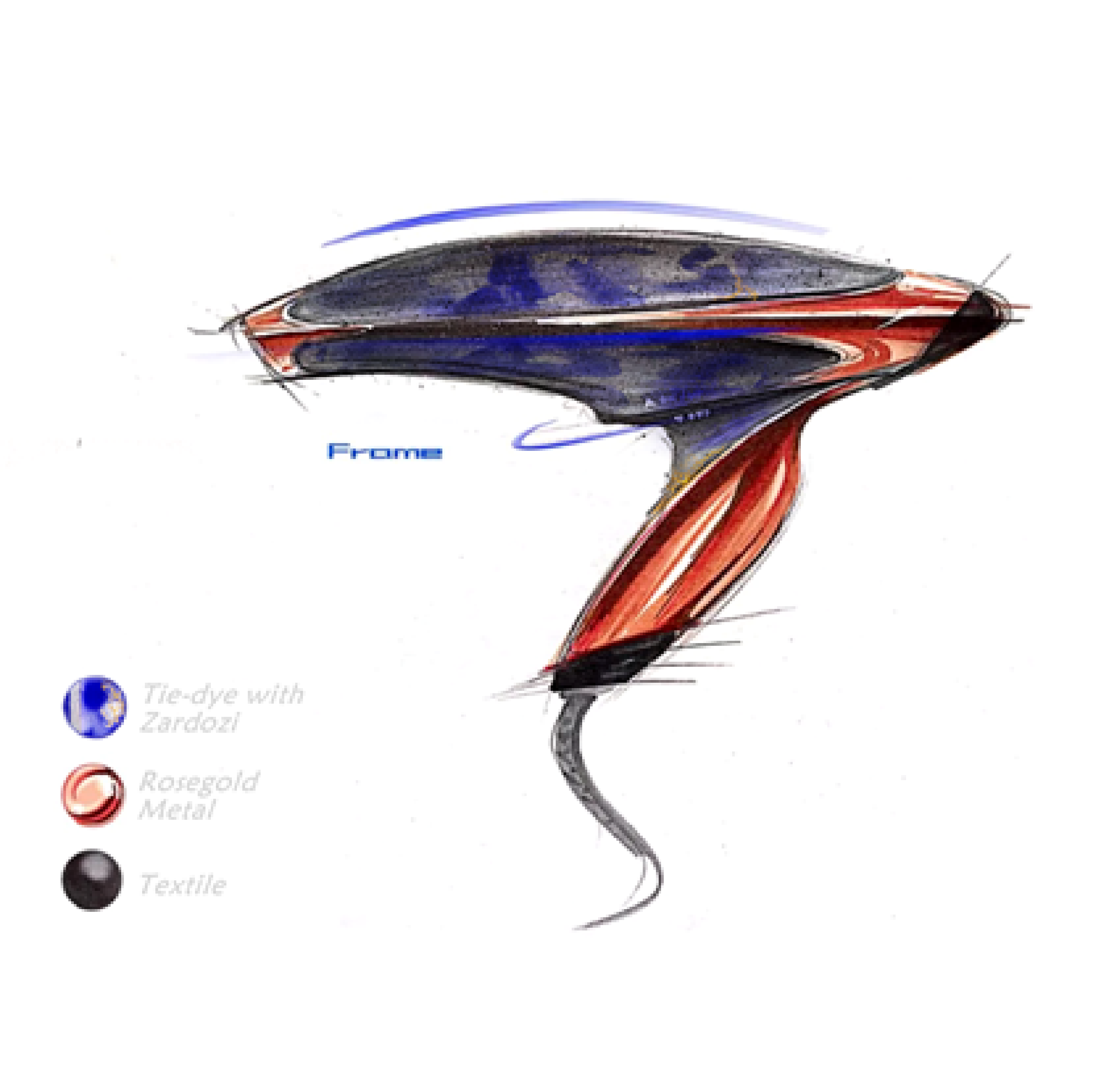}
	\caption{The sketch of hair dryer inspired by "Tie-dye" and "Zardozi"} 
	\label{fig:appearance}
\end{figure*}

\begin{figure*}[t!]
	\centering
	\includegraphics[width=0.4\textwidth]{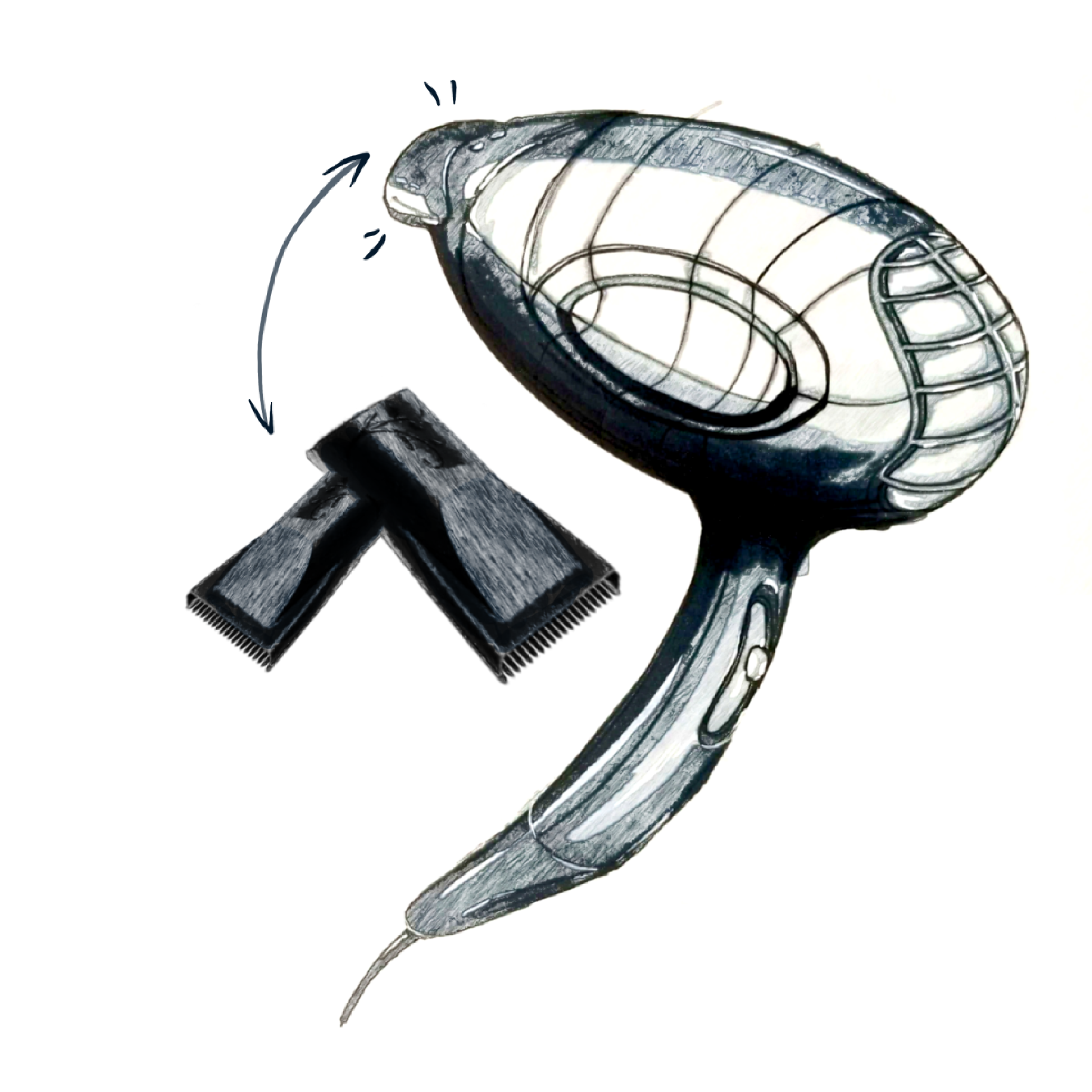}
	\caption{The sketch of hair dryer inspired by "Comb"}
	\label{fig:function}
\end{figure*}

In particular, we use  Figure \ref{fig:appearance} and \ref{fig:function} are the designs produced with the inspiration links "Hair dryer" – "Comb" – "Hairstyle" - "Tie-dye" – "Zardozi".
 
In particular, two ideas were generated during the designer's manipulation with WikiLink. The first design, as shown in Figure \ref{fig:appearance}, is an appearance design inspired by "Tie-dye" and "Zardozi". The existing hair dryers in the market is mostly in pure color with a smooth or frosted plastic shell. "Tie-dye", the characteristic of the Bai nationality, has special patterns which are uneven in-depth and rich in layers, and overcomes the rigidity of pure color. "Zardozi", a traditional  Chinese craft, has a delicate touch feeling compared with plastic material. Thus "Tie-dye" and "Zardozi" inspire the designer to integrate traditional Chinese cultural elements into the design of hair dryer to increase cultural connotation. The second design (Figure \ref{fig:function}) is functional and inspired by "Hairstyle" and "Comb" in the design inspiration link. The idea is to design the replaceable hair dryer nozzle with the features of "Comb" so that users can comb their hair conveniently while drying the hair without searching it in a hurry.

\section{Conclusion}
In this research, firstly, a semantic network for design innovation is constructed. Wikipedia is applied as the data source for the semantic network. During the construction, the Wikipedia items are regarded as the nodes, the interlinks between the items on the same page are regarded as the directly connected relationship (edges) between nodes. The evaluation result indicate that the network contains information from a wide range of fields and expands the data to a new boundary. Secondly, instead of simply one type of weight, a combined weight is introduced for the relationship in the semantic network. The combined weight fuses the statistical relationship and semantic relationship which better captures the implicit connection between concepts for design innovation. Four algorithms are further developed to retrieve relevant knowledge concepts and relationships with different levels and manners. Thirdly, the constructed semantic network for design innovation is further developed as a tool, called WikiLink. An evaluation and demonstration for WikiLink are conducted subsequently. Compared with other benchmarks, with the fusion of semantic meaning weight and statistical weight, WikiLink can well balance the breadth and depth in exploring knowledge for design innovation. A design case is conducted to demonstrate the process of how WikiLink can facilitate idea generation. The results indicate that WikiLink can serve as a design ideation tool for design innovation.

The study leaves some space for future research though it provides a functional panel for practical use. The weight strength fusing two types of weight is one of the main contributions in this research, but it only shows the numerical value and lacks of explicit semantic meaning describing the relationship between two concepts. Thus a semantic description is expected to be added to the edges in WikiLink and provide richer information for design innovation. Besides, the network visualization of WikiLink is currently designed on a two-dimensional scale, which might cause information explosion when the retrieved network keeps growing. A three-dimensional scale network along with other information visualization techniques could be a solution and provide a more dynamic way for users to explore information and obtain inspiration more effectively.

\bibliographystyle{model2-names}
%

\bibliography{cas-refs}





\end{document}